\pdfoutput=1
\documentclass[11pt]{article}

\usepackage[preprint]{acl}

\usepackage{times}
\usepackage{latexsym}

\usepackage[T1]{fontenc}
\usepackage[utf8]{inputenc}

\usepackage{placeins}

\usepackage{microtype}

\usepackage{inconsolata}

\usepackage{graphicx}

\title{Can LLMs Interpret and Leverage Structured Linguistic Representations? A Case Study with AMRs}

\author{Ankush Raut$^1$ \;\; Xiaofeng Zhu$^2$ \;\; Maria Leonor Pacheco$^1$ \\
$^1$University of Colorado Boulder \;\; $^2$Northwestern University\\
$^1$\texttt{\{ankush.raut, maria.pacheco\}@colorado.edu} \;\; $^2$\texttt{xiaofengzhu2013@u.northwestern.edu}}

\usepackage{amsmath}

\begin{document}
\maketitle
\begin{abstract}
This paper evaluates the ability of Large Language Models (LLMs) to leverage contextual information in the form of structured linguistic representations. Specifically, we examine the impact of encoding both short and long contexts using Abstract Meaning Representation (AMR) structures across a diverse set of language tasks. We perform our analysis using 8-bit quantized and instruction-tuned versions of Llama 3.1 (8B), Phi-3, and Mistral 7B. Our results indicate that, for tasks involving short contexts, augmenting the prompt with the AMR of the original language context often degrades the performance of the underlying LLM. However, for tasks that involve long contexts, such as dialogue summarization in the SAMSum dataset, this enhancement improves LLM performance, for example, by increasing the zero-shot cosine similarity score of Llama 3.1 from 66\% to 76\%. This improvement is more evident in the newer and larger LLMs, but does not extend to the older or smaller ones. In addition, we observe that LLMs can effectively reconstruct the original text from a linearized AMR, achieving a cosine similarity of 81\% in the best-case scenario. %This paper also serves as a guide for analyzing how LLMs interpret other representation structures, such as Knowledge Graphs and Discourse Representation Structures.
\end{abstract}

\section{Introduction}
\label{Introduction}

In recent years, Large Language Models (LLMs) have achieved remarkable success in numerous natural language processing (NLP) tasks, such as machine translation, summarization, and both single- and multi-hop question answering. However, a critical challenge remains: assessing their ability to interpret and utilize meaning from structured representations that encode language in concise and abstract forms. Structured semantic representations of text, such as Abstract Meaning Representation (AMR) structures and Discourse Representation Structures (DRS), have consistently proven effective for reasoning with high-level semantics in text and enhancing performance in challenging structure-aware NLP tasks, particularly those involving long contexts. For instance, prior work has demonstrated that encoding AMR structures via text-graph attention can improve long-dialogue summarization performance \cite{hua-etal-2023-improving}.

In this paper, we systematically examine the ability of several prominent LLMs to interpret AMRs. In addition, we study how AMR-augmented prompting (See Fig. \ref{fig:full_ex}) and AMR-only prompting (See Fig. \ref{fig:snli_oamr_prompt}) affect model performance on downstream tasks, that require a deep understanding of context, compared to context-only, i.e., language-only prompting. Unlike previous work, which primarily focuses on developing ad-hoc neural architectures to either improve text regeneration from AMRs \cite{zhu-etal-2019-modeling} or enhance performance on downstream tasks \cite{hua-etal-2023-improving, yang2024emphasisingstructuredinformationintegrating}, this study evaluates the ability of LLMs to directly interpret linearized (flattended) AMRs. Here, AMR-augmented prompting refers to including the linearized AMR of the context in the prompt to evaluate the model's ability to improve its understanding of the original context with support from the AMR. AMR-only prompting refers to providing the LLM with only the linearized AMR, without the original language context, to assess the model's ability to infer meaning directly from these semantic representations. Our key contributions are as follows:

1. We systematically evaluate the instruction-tuned versions of prominent LLMs on their ability to directly leverage linearized AMRs in prompts for both short- and long-context tasks. These tasks include AMR-to-text generation, single-hop reasoning, 2-hop reasoning, dialogue summarization, natural language inference (NLI), and document-level natural language inference (DocNLI).

2. We demonstrate that while AMR integration tends to degrade performance in short-context tasks, it can enhance performance in long-context tasks, an observation that is more profound when using larger and more recently developed LLMs. Additionally, in some downstream tasks, LLMs can work exclusively with linearized AMRs (AMR-only prompting) to achieve reasonable performance with few-shot prompting.

3. We show that LLMs can effectively reconstruct the original context from linearized AMRs, achieving 81\% cosine similarity in the best-case scenario, highlighting their ability to understand context from AMRs.

%4. We provide a guide for comprehensively analyzing other structured semantic representations, such as Knowledge Graphs and Discourse Representation Structures, in terms of how they are interpreted and utilized by LLMs.

This work provides valuable insights into leveraging AMRs to enhance LLM performance on downstream language tasks, while also identifying key challenges. We hope that our work will serve as a guide for comprehensively analyzing other structured representations, such as Knowledge Graphs and Discourse Representation Structures, with respect to how they can be interpreted and utilized by LLMs.

\section{Related Work}

Existing studies have explored various methods for generating text from language and knowledge representation structures, usually by modifying or improving underlying neural architectures. For instance, approaches like text-graph attention \cite{hua-etal-2023-improving} and JointGT \cite{ke2021jointgtgraphtextjointrepresentation} alter the transformer architecture using structured cross-attention to better account for graph properties. Prior studies have also demonstrated enhanced open-domain dialogue evaluation by fusing graph-encoded AMRs into LLMs via gating mechanism \cite{yang2024emphasisingstructuredinformationintegrating}. Others, such as text generation from knowledge graphs using graph transformers \cite{koncelkedziorski2022textgenerationknowledgegraphs}, use graphical representations directly, alleviating the need for graph linearization. These studies have aimed to enhance the quality of generated text using AMRs as input through structure-aware methods, semantic aggregation, and other heuristics.

However, all of these approaches exhibit the key limitation that the proposed work seeks to address. Namely, they focus on modifying model architectures or introducing new architectural components, such as joint graph-text representations and heuristic-based conditioning, which can increase complexity and is less amenable to generalization across domains, tasks, and structures. In contrast, we directly evaluate the inherent capability of existing, general-purpose LLMs to interpret structured representations like AMR. Moreover, while most prior studies largely sought to improve the quality of generated text \cite{Dong2014NaturalLG}, the proposed work %is based on understanding how LLMs infer from these representations without undergoing structural alterations and 
evaluates the opportunities of AMRs to represent context for a wide range of downstream language tasks. 

\section{Methodology}
\label{Methodology}

This section describes AMRs, how they are extracted from text, and the pipeline for studying their efficacy in language understanding and reasoning tasks. 

\subsection{Abstract Meaning Representation}

\begin{figure}
    \centering
    \includegraphics[width=1\linewidth]{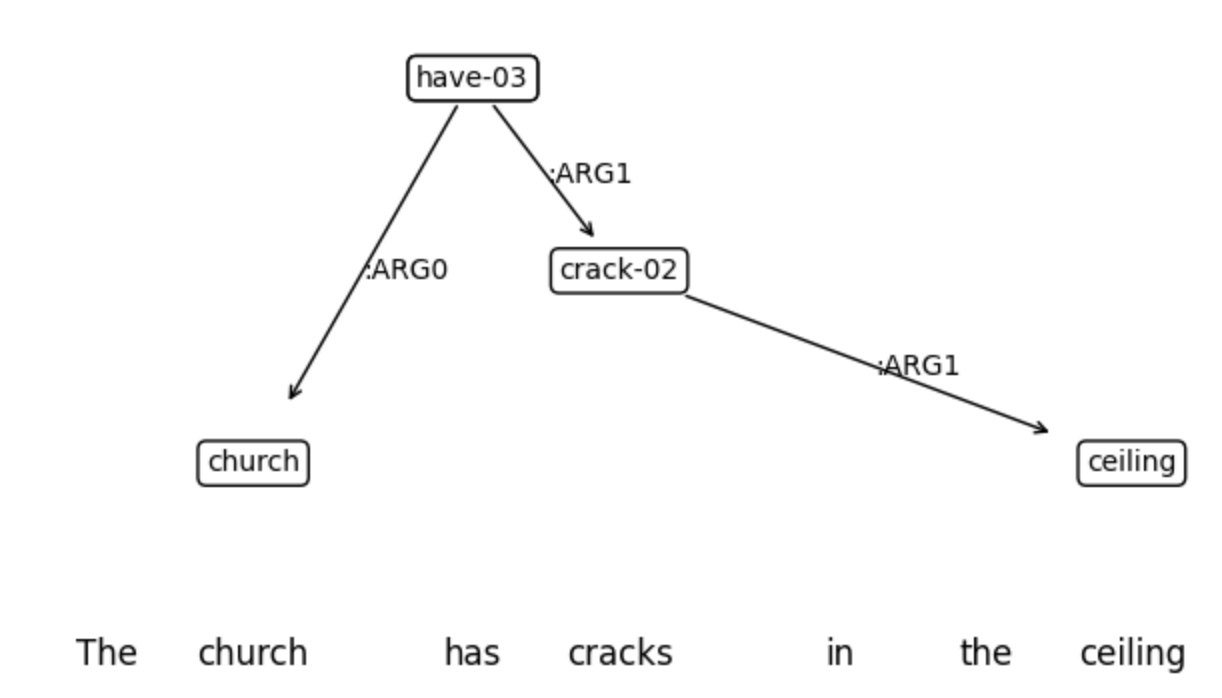}
    \caption{An AMR tree for a hypothesis from the SNLI dataset, \textit{The church has cracks in the ceiling}, extracted using the AMR3-structbart-L model via IBM's transition neural parser.}
    \label{fig:amr_tree}
\end{figure}

An Abstract Meaning Representation (AMR) \cite{langkilde-knight-1998-generation-exploits} is a labeled representation of sentences as rooted, directed graphs (See Fig. \ref{fig:amr_tree}) that capture the semantic meaning of a sentence by abstracting away from its surface syntax. The most basic AMR takes the form \textit{(label / concept)}, e.g. \[ (m1 \ / \ |dog<canid|) \]
The slash (/) is shorthand for a type (or instance) feature, and in logical notation, this AMR might be written as \textit{instance}(m1, dog). This AMR can represent "the dog," "the dogs," "a dog," or simply "dog." A concept can be modified using keywords:\[
\begin{array}{l}
    \bigl( m1 \,/\, |dog<canid| \\
    \ \ \ \ \ \ \ \ \ :quant \ plural \bigr)
\end{array}
\]
This specification refines the meaning to "the dogs" or simply "dogs." AMR prioritizes the underlying semantic meaning conveyed by a sentence rather than the specific words or syntactic structures used.

% \begin{figure}
%     \centering
%     \includegraphics[width=1\linewidth]{amr_pic.png}
%     \caption{AMR structure for \textit{The boy wanted to believe the girl.}}
%     \label{fig:amr}
% \end{figure}

\begin{figure}
    \centering
    \includegraphics[width=1\linewidth]{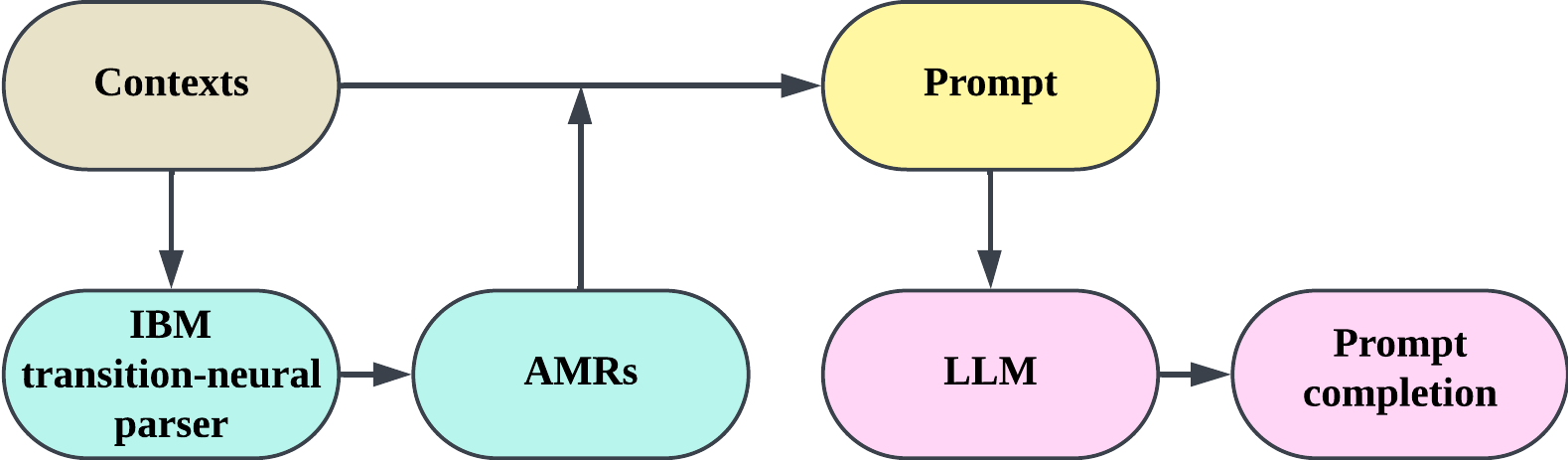}
    \caption{Process flow for the analysis tasks.}
    \label{fig:process-flow}
\end{figure}

\begin{figure}
    \centering
    \includegraphics[width=0.8\linewidth]{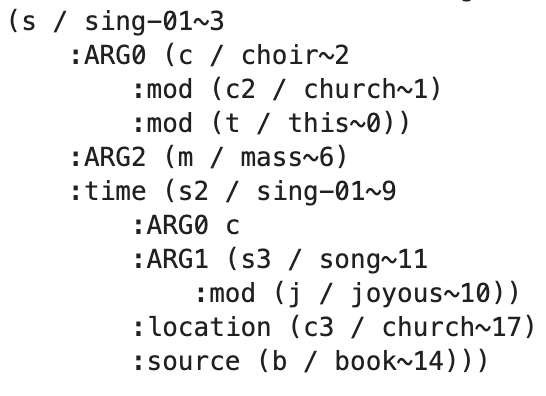}
    \caption{Linearized AMR for a premise from the SNLI dataset, \textit{This church choir sings to the masses as they sing joyous songs from the book at a church}, extracted using the AMR3-structbart-L model via IBM's transition neural parser.}
    \label{fig:lin-amr}
\end{figure}

In this paper, we use the AMR Annotation Release 3.0 dataset (LDC2020T02), developed by the Linguistic Data Consortium (LDC), SDL/Language Weaver, Inc., the Computational Language and Educational Research (CLEAR) group at the University of Colorado, and the Information Sciences Institute at the University of Southern California. This dataset contains a semantic tree-bank of over 59,255 English natural language sentences from sources including broadcast conversations, news-wire articles, weblogs, web discussion forums, fiction, and web text. There is a possibility that the AMRs in this dataset were exposed in the pre-training of many LLMs, which is why we also employ the AMR3-structbart-L and doc-sen-conll-amr-seed42 models via IBM's transition-based neural parser \cite{drozdov2022inducingusingalignmentstransitionbased} to parse contexts from various datasets into document/multi-sentence-level AMR structures for further analysis. Linearized representations of AMRs (See Fig. \ref{fig:lin-amr}) were fed to the LLMs. Fig. \ref{fig:process-flow} illustrates a general process flow for all the analysis tasks in this work.

\subsection{Tasks}
\label{Tasks}

All tasks were approached using zero-shot, 3-shot, and 5-shot prompting. For all tasks except context regeneration, prompting was done in 3 ways: context-only, AMR-augmented, and AMR-only.

\subsubsection{Context Regeneration (AMR-to-text)} 

\begin{figure}
    \centering
    \includegraphics[width=0.7\linewidth]{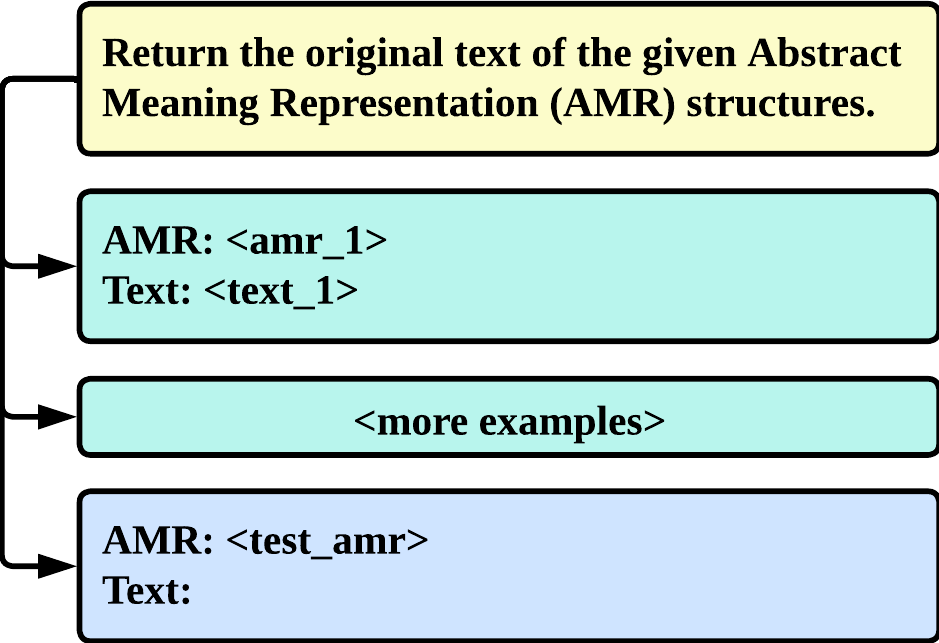}
    \caption{Prompt for regenerating text from AMRs.}
    \label{fig:regen-prompt}
\end{figure}

In this task, we evaluate how well LLMs can regenerate the original context given its linearized AMR. %Zero-shot prompting yields erratic results, as the LLMs require some examples to understand the annotations used in the linearized AMRs. 
Regeneration is conducted using the LDC2020T02 dataset. Fig. \ref{fig:regen-prompt} illustrates the prompting strategy for context regeneration.

\subsubsection{Question-Answering (QA)} 

\begin{figure}[t]
    \centering
    \includegraphics[width=1\linewidth]{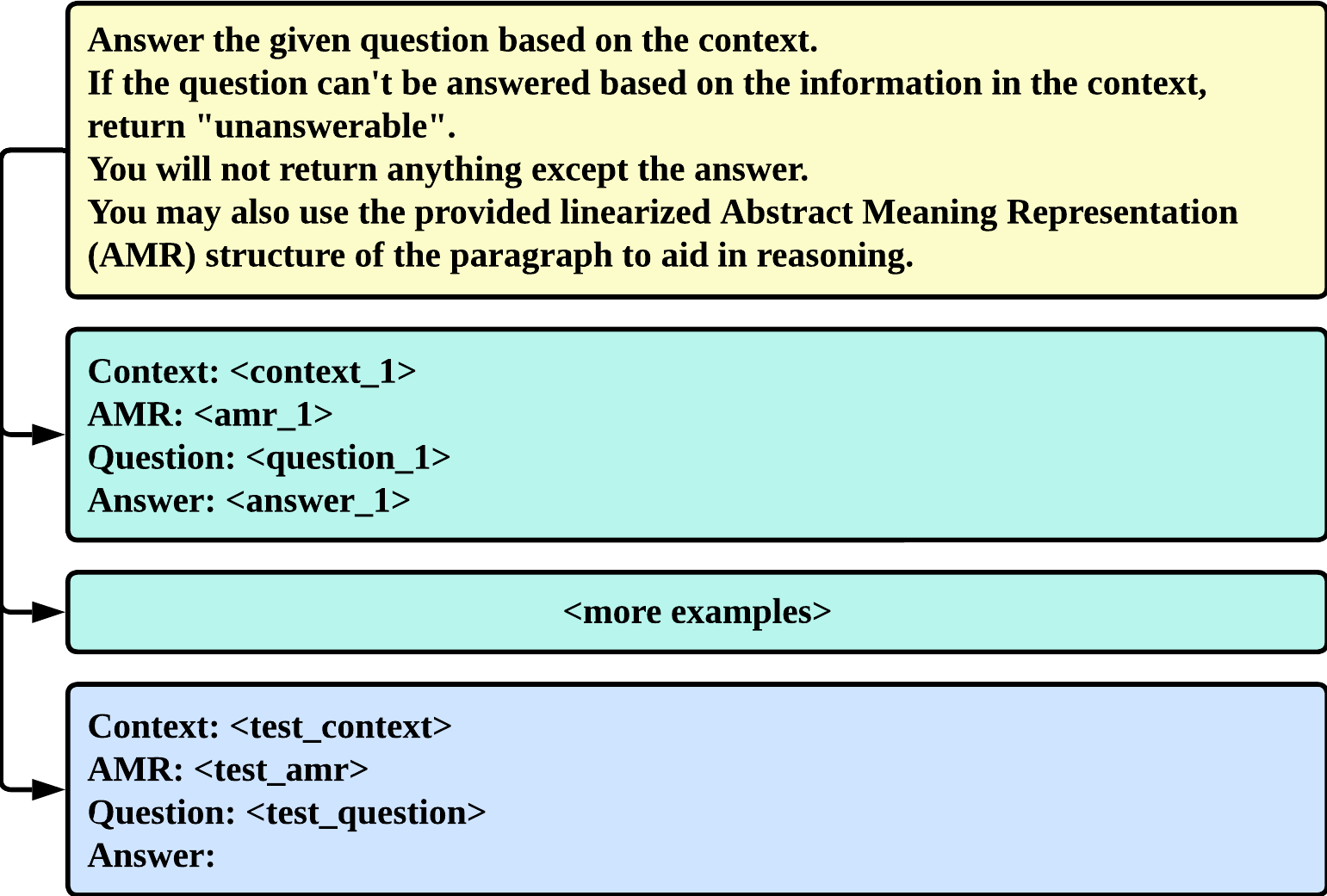}
    \caption{AMR-augmented prompt for QA on the SQuAD 2.0 dataset. For context-only prompting (Fig. \ref{fig:squad_raw_prompt}), the final statement from the canary-colored instruction block is removed, along with the AMRs from the examples. For AMR-only prompting (Fig. \ref{fig:squad_oamr_prompt}), the instruction is modified so that the LLM considers the AMR as the context for reasoning, and the original context is removed from the examples.}
    \label{fig:squad}
\end{figure}

\begin{figure}[t]
    \centering
    \includegraphics[width=1\linewidth]{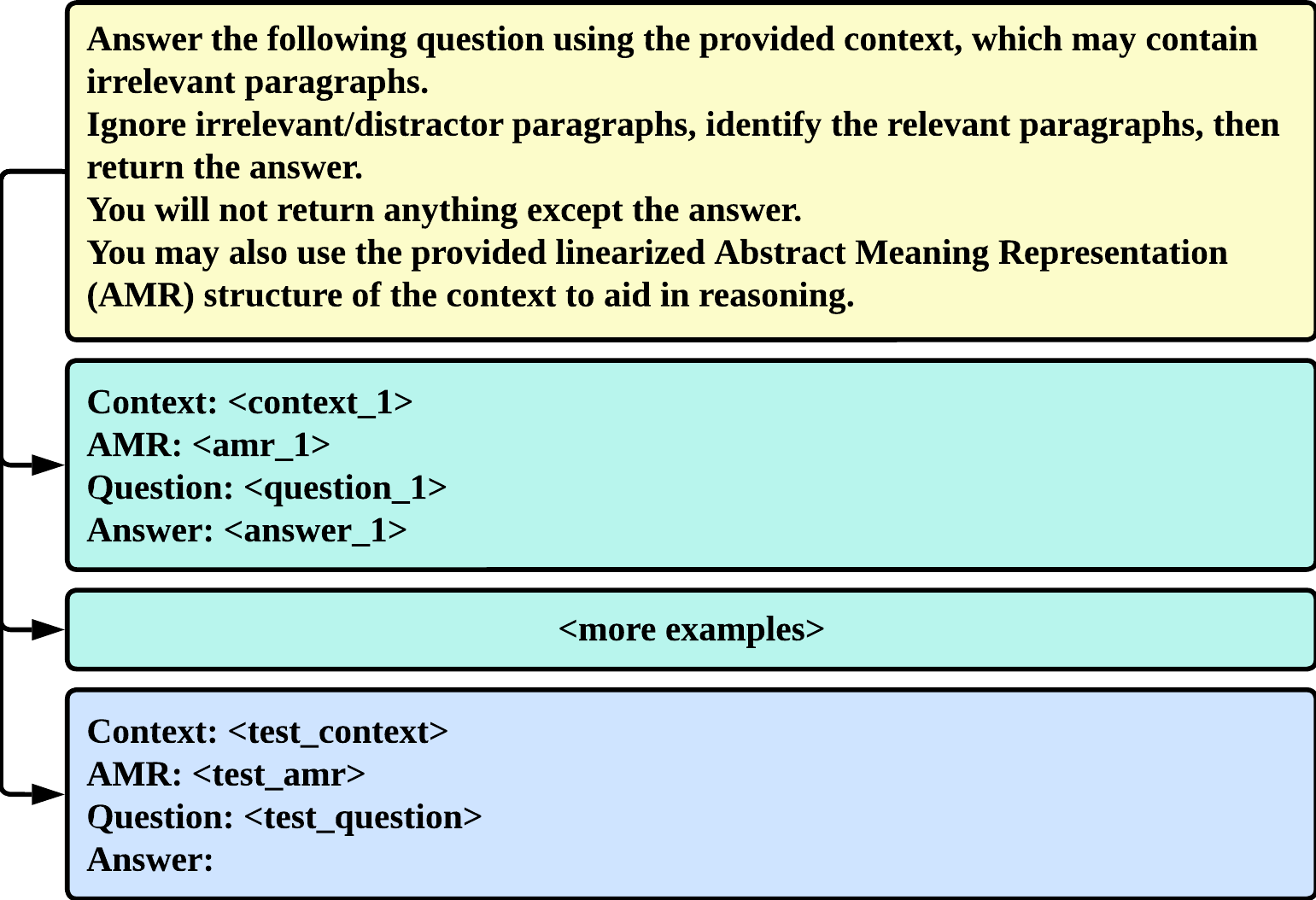}
    \caption{AMR-augmented prompt for QA on HotpotQA dataset. The context-only (Fig. \ref{fig:hotpot_raw_prompt}) and AMR-only (Fig. \ref{fig:hotpot_oamr_prompt}) prompting strategies are similar to those used in SQuAD 2.0 reasoning.}
    \label{fig:hotpot}
\end{figure}

In this task, we evaluate the effectiveness of AMRs in enhancing the single-hop and 2-hop reasoning abilities of LLMs. For single-hop QA, we report performance on the SQuAD 2.0 dataset \cite{rajpurkar-etal-2018-know}. Fig. \ref{fig:squad} illustrates the AMR-augmented prompt used for QA on the SQuAD 2.0 dataset. For 2-hop reasoning, we report performance on the HotpotQA dataset \cite{yang2018hotpotqadatasetdiverseexplainable}, which features much longer contexts, as each question in that dataset is accompanied by 10 documents (2 relevant, 8 irrelevant). Since most of these documents are distractors, the prompt includes a specific instruction indicating that some documents in the context may be irrelevant to the question. Fig. \ref{fig:hotpot} illustrates the AMR-augmented prompt used for QA on the HotpotQA dataset.

\subsubsection{Summarization} 

\begin{figure}[t]
    \centering
    \includegraphics[width=1\linewidth]{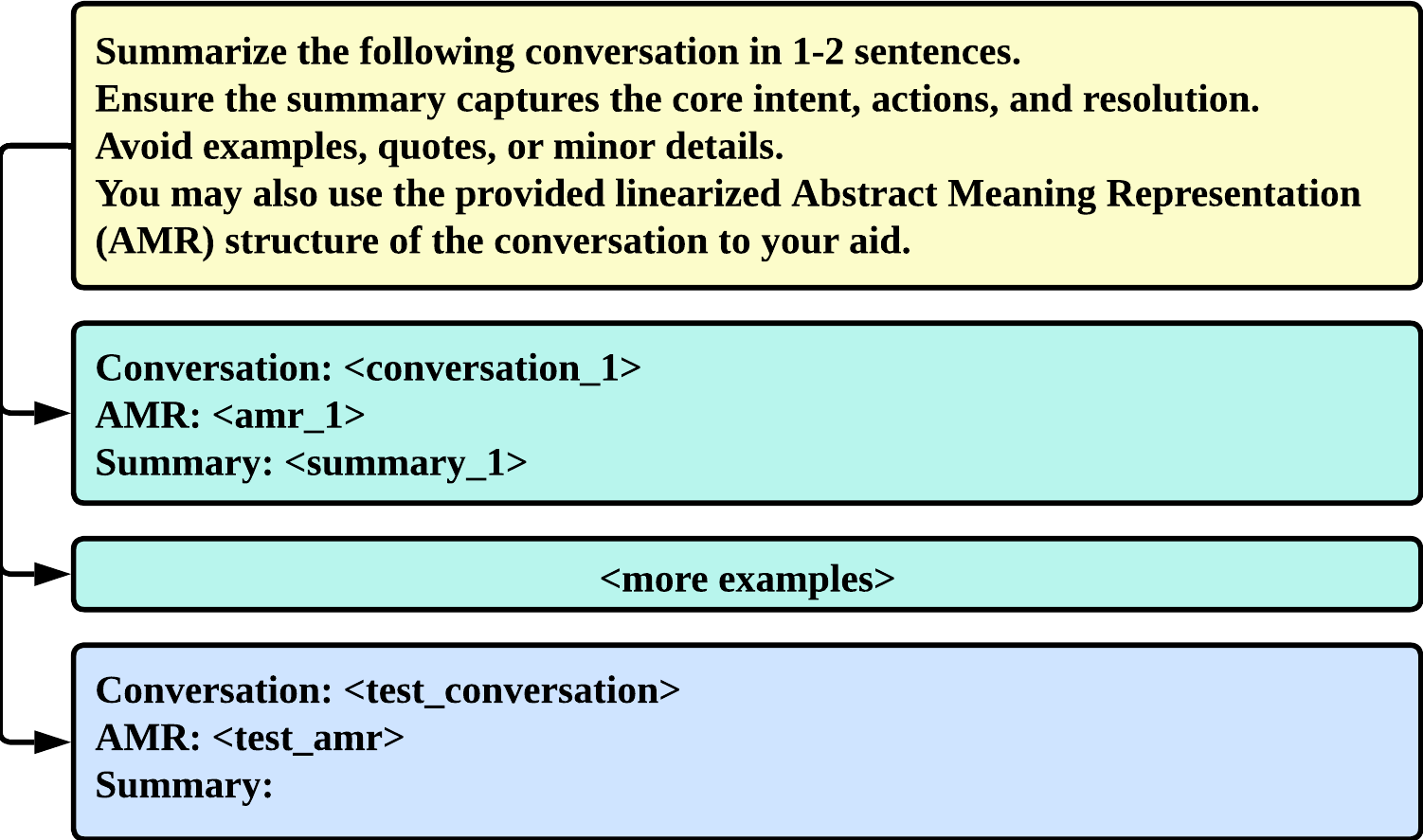}
    \caption{AMR-augmented prompt for summarizing SAMSum dataset conversations. The context-only (Fig. \ref{fig:samsum_raw_prompt}) and AMR-only (Fig. \ref{fig:samsum_oamr_prompt}) prompting strategies are similar to those used in SQuAD 2.0 reasoning.}
    \label{fig:samsum}
\end{figure}

In this task, we evaluate the effectiveness of AMRs in enhancing the dialogue summarization capabilities of LLMs, comparing these summaries to the gold (expert human-generated) summaries in the dataset. Fig. \ref{fig:samsum} illustrates the AMR-augmented prompt for summarizing conversations in the SAMSum dataset \cite{gliwa-etal-2019-samsum}.

\subsubsection{Natural Language Inference (NLI)} 

\begin{figure}[t]
    \centering
    \includegraphics[width=1\linewidth]{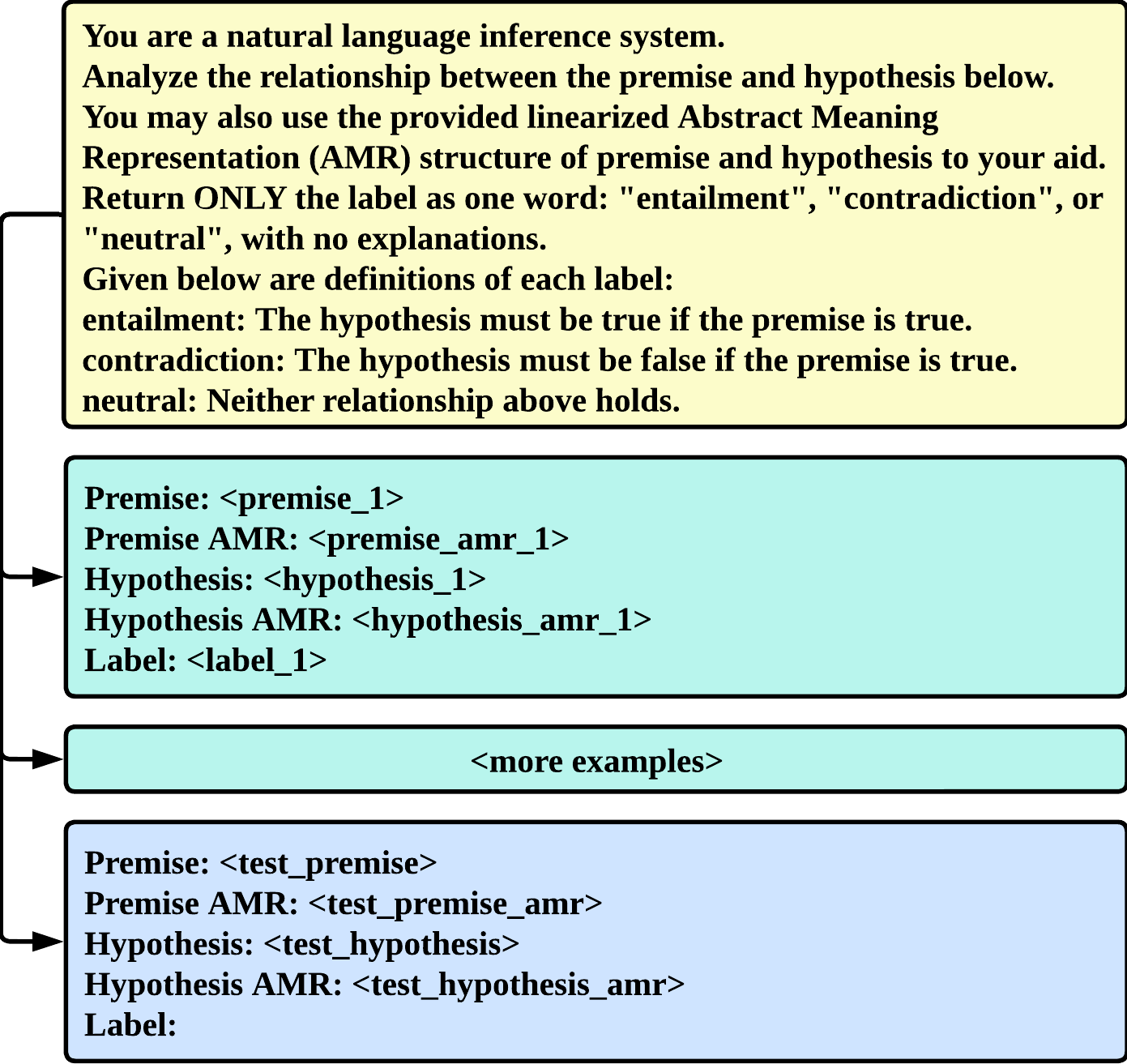}
    \caption{AMR-augmented prompt for natural language inference on SNLI dataset. The context-only (Fig. \ref{fig:snli_raw_prompt}) and AMR-only (Fig. \ref{fig:snli_oamr_prompt}) prompting strategies are similar to those used in SQuAD 2.0 reasoning. The 3-shot examples for SNLI included 1 example of each label, while the 5-shot examples included 2 examples each of entailment and contradiction, and 1 neutral example. For DocNLI, due to an imbalance in the test set (with more examples of contradiction than entailment), we included one more example of contradiction than entailment in the few-shot prompts.}
    \label{fig:snli_prompt}
\end{figure}

In this task, we evaluate the effectiveness of AMRs in enhancing the ability of LLMs to understand and reason about natural language, specifically, to determine whether a given claim is supported by sentences in an evidence document. This involves identifying if there is an entailment between the claim (hypothesis) and the supporting evidence (premise) or if a contradiction exists between them. We use the SNLI dataset \cite{bowman2015largeannotatedcorpuslearning}, composed of pairs of short statements, and the DocNLI dataset \cite{yin-etal-2021-docnli}, where premises can span multiple sentences and documents. Fig. \ref{fig:snli_prompt} illustrates the AMR-augmented prompt for natural language inference on the SNLI dataset. A similar prompt was used for DocNLI, with the removal of the neutral label from the instruction to align with the binary labels in DocNLI.

Fig. \ref{fig:full_ex} in the Appendix illustrates an example of a prompt-completion for the 3-shot SNLI entailment prediction task.

\section{Experimental Settings}

Our analysis pipeline involves converting text into AMR structures and then prompting LLMs to perform tasks using those AMRs, either via AMR-augmented or AMR-only prompting. The baselines for these tasks are simply the language-only (i.e., context-only) prompting results. In this section, we will describe the experimental setup for the tasks outlined in Section \ref{Tasks}.

\subsection{Datasets and Splits}
In addition to using the LDC2020T02 dataset, which contains AMRs of small contexts (sentence-level AMRs), we also parse contexts from the SQuAD 2.0, HotpotQA, SAMSum, SNLI, and DocNLI datasets into document-level (multi-sentence) AMRs using the AMR3-structbart-L and doc-sen-conll-amr-seed42 models via IBM's transition-based neural parser. Note: a document-level AMR corresponds to a single, continuous document. For instance, each question in the HotpotQA dataset provides up to 10 documents, which may or may not be related. This setup results in 10 document-level AMRs.

We use the test splits of LDC2020T02, SAMSum, SNLI, and DocNLI, and the validation splits of SQuAD 2.0 and HotpotQA for inference. The few-shot examples are curated from the training splits of these datasets. We also conduct a fine-tuning experiment for SAMSum summarization using its training and validation splits.

\subsection{Models}

For all tasks, we generate responses truncated at the first occurrence of a newline character using 8-bit quantized versions of Llama-3.1-8B-Instruct (Llama3.1) \cite{grattafiori2024llama3herdmodels}, Phi-3-mini-128k-instruct (Phi3) \cite{abdin2024phi3technicalreporthighly}, and Mistral-7B-Instruct-v0.1 (Mistral) \cite{jiang2023mistral7b}. Table \ref{tab:mapping} outlines the mapping of models to their respective tasks.

We also experiment with rank-32 LoRA \cite{hu2021loralowrankadaptationlarge} fine-tuning of the 8-bit quantized Llama3.1 model for the SAMSum summarization task. We then compare the performance of this fine-tuned model against few-shot summarization.

\subsection{Evaluation Metrics}

Listed below are the evaluation metrics used for each task, according to their output types.

\paragraph{AMR-to-text and Summarization} For these tasks, we report the ROUGE-1, ROUGE-2, ROUGE-L, BLEU, and cosine similarity scores comparing the expected response to the generated response averaged across all samples. The cosine similarity between the all-MiniLM-L6-V2 \cite{wang2020minilmdeepselfattentiondistillation} embeddings of the expected response and the generated response averaged across all samples is reported as the cosine similarity score.

\paragraph{Question Answering} We report F1-score for all question answering tasks. 
The F1 is calculated for each sample based on the number of common tokens between the expected and generated responses, then averaged across all samples. This is operationalized as:
\[
\text{Precision} = \frac{\text{count of common tokens}}{\text{count of tokens in generated response}}
\]
\[
\text{Recall} = \frac{\text{count of common tokens}}{\text{count of tokens in generated response}}
\]
\[
F1 = 2 \times \frac{\text{Precision} \times \text{Recall}}{\text{Precision} + \text{Recall}}
\]
We also report cosine similarity score for this task. For SQuAD 2.0, we select the best possible match from the available answer choices for scoring.

\paragraph{Natural Language Inference} For SNLI and DocNLI, we report the macro F1-score of the generated answers compared to the labels; \textit{contradiction, neutral, entailment} for SNLI and \textit{contradiction, entailment} for DocNLI. The macro F1-score averages the F1-scores across all classes, thereby accounting for label imbalance and providing a more comprehensive measure of classification performance as compared to accuracy.

\begin{table}[t]
    \begin{tabular}{|c|c|c|c|c|}
    \hline
        Task & Llama3.1& Phi3 & Mistral \\
        \hline
        Regeneration & Yes & Yes & Yes \\
        \hline
        Summarization & Yes & Yes & Yes \\
        \hline
        1-hop QA & Yes & Yes & Yes \\
        \hline
        2-hop QA & Yes & No & No \\
        \hline
        SNLI & Yes & Yes & Yes \\
        \hline
        DocNLI & Yes & No & No \\
        \hline
    \end{tabular}
    \caption{Mapping models to the tasks for which they will be used. Only Llama3.1 is employed for 2-hop QA and DocNLI due to its ability to handle long contexts efficiently, which the other models lack.}
    \label{tab:mapping}
\end{table}

\section{Results}

All scores presented in the charts and tables are expressed as percentages. For each task, except for AMR-to-text, we present the results of the best performing model in the main paper and report the remaining results in the Appendix.

For all the zero-shot experiments except AMR-to-text, 5 different transformer seeds were used to compute 90\% confidence intervals. Similar performance was recorded across all seeds for zero-shot experiments, resulting in narrow confidence intervals. For all the few-shot experiments except AMR-to-text, 5 distinct sets of examples were curated from the training set to determine 90\% confidence intervals. All the scores presented in the results and tables are averaged across different seeds and sets, if applicable.

\subsection{LDC2020T02}

\begin{figure}
    \centering
    \includegraphics[width=1\linewidth]{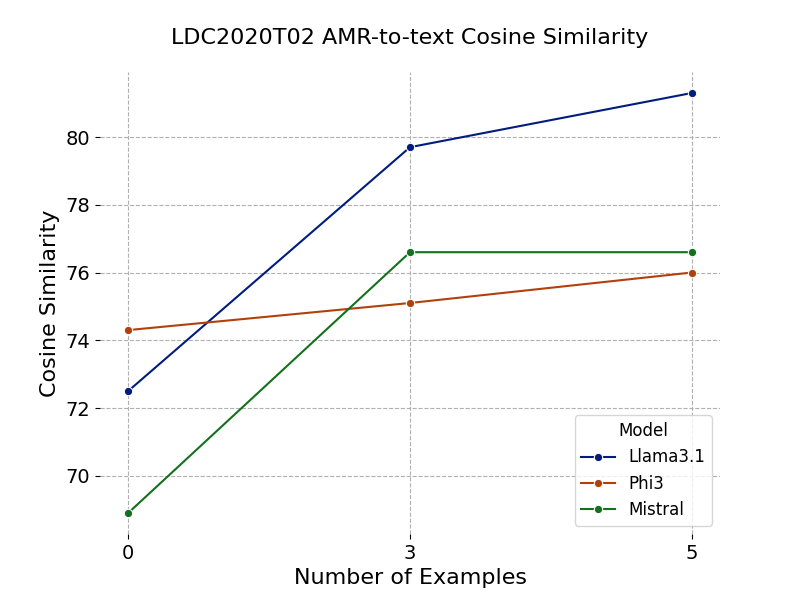}
    \caption{LDC2020T02 AMR-to-text Cosine Similarity.}
    \label{fig:ldc_result}
\end{figure}

Fig. \ref{fig:ldc_result} illustrates that Llama3.1 outperforms Phi3 and Mistral in regenerating the original contexts from the linearized LDC2020T02 AMRs using few-shot prompting. While Phi3 achieves the best zero-shot regeneration cosine similarity score (74\%), it becomes the worst performer with few-shot prompting. With 5-shot prompting, Llama3.1 attains a cosine similarity score of 81\%. Using more than 3 in-context learning examples yields diminishing returns, as indicated by the marginal improvement when transitioning from 3-shot to 5-shot prompting. Detailed results for this task, including additional regeneration metrics, are provided in Table \ref{tab:ldc} in the Appendix.

\subsection{SAMSum}

\begin{figure}
    \centering
    \includegraphics[width=1\linewidth]{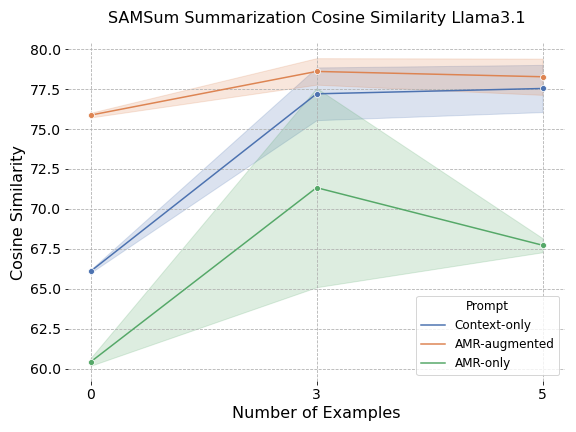}
    \caption{Llama3.1 SAMSum summarization Cosine Similarity with 90\% confidence intervals.}
    \label{fig:samsum_cosine}
\end{figure}

Figure~\ref{fig:samsum_cosine} illustrates how Llama3.1's zero-shot summarization cosine similarity score improves from 66\% to 76\% with AMR-augmented prompting, compared to context-only prompting. In the few-shot setting, although the confidence intervals for context-only and AMR-augmented prompting overlap, both the average and maximum cosine similarity scores are higher for AMR-augmented prompting than for other prompting strategies. This provides strong evidence that AMRs can enhance the ability of LLMs, particularly larger and more recent models, to infer information from long contexts, especially lengthy dialogues, and to retain key pieces of information. However, increasing the number of few-shot examples beyond 3 yields diminishing returns for both context-only and AMR-augmented prompting, and it even degrades performance with AMR-only prompting.

While Llama3.1's performance improved with AMR augmentation, Phi3 and Mistral did not exhibit similar gains (Table~\ref{tab:samsum}). Mistral was the best performing model for this task by a small margin, despite being worse than Llama3.1 in all the other tasks. Rank-32 LoRA fine-tuning of Llama3.1 achieved a cosine similarity score of 75\% with a context-only input prompt, which increased to 76\% with an AMR-augmented input prompt. This indicates that, although LoRA fine-tuning of Llama3.1 was not any better than few-shot prompting for SAMSum summarization, it benefited marginally from AMR augmentation.

\subsection{SQuAD 2.0}

\begin{figure}
    \centering
    \includegraphics[width=1\linewidth]{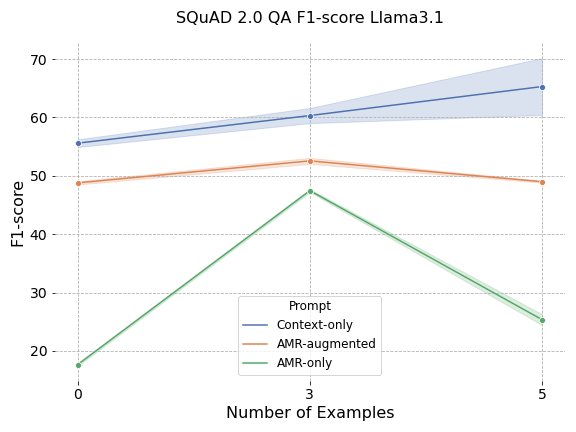}
    \caption{Llama3.1 SQuAD 2.0 QA F1-score with 90\% confidence intervals.}
    \label{fig:sq_f1}
\end{figure}

Fig. \ref{fig:sq_f1} illustrates that Llama3.1's single-hop reasoning F1-score declines with AMR-augmented prompting compared to context-only prompting (e.g., from 59\% to 52\% with 3-shot prompting). However, both Llama3.1 (Fig. \ref{fig:sq_f1}) and Mistral (Table \ref{tab:squad}) demonstrate a remarkable understanding of AMRs in 3-shot single-hop reasoning, as indicated by how closely their 3-shot AMR-only performance approaches the best possible 3-shot performance for this task. Including over 3 examples in the prompt results in only marginal performance improvements with context-only prompting. However, with AMR-augmented and AMR-only prompting, performance deteriorates as the number of few-shot examples increases to 5. For instance, Llama3.1 achieved a 48\% F1 score with AMR-only 3-shot prompting, which dropped significantly to 26\% with AMR-only 5-shot prompting. Detailed results for this task, including results from other models, are provided in Table \ref{tab:squad} in the Appendix.

\subsection{HotpotQA}

\begin{figure}
    \centering
    \includegraphics[width=1\linewidth]{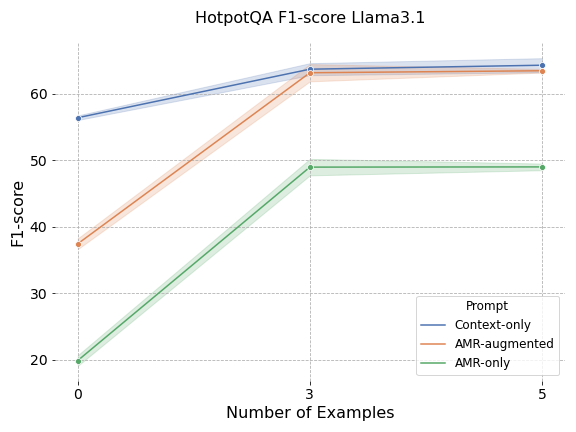}
    \caption{Llama3.1 2-hop HotpotQA F1-score with 90\% confidence intervals.}
    \label{fig:hotpot_f1}
\end{figure}

Fig. \ref{fig:hotpot_f1} shows that AMR-augmented prompting in Llama3.1 does not outperform context-only prompting for 2-hop reasoning on the HotpotQA dataset. Additionally, across all prompting scenarios, increasing from 3-shot to 5-shot prompting has insignificant impact on performance. Although this is a long-context task, the individual documents within the context are small, resulting in multiple AMRs of these small documents being stacked together in the input prompt. This explains why the observations regarding the effectiveness of AMRs here differ from those in the SAMSum dataset, where AMRs were derived from long conversations and proved more effective in improving task performance.

\subsection{SNLI}

\begin{figure}
    \centering
    \includegraphics[width=1\linewidth]{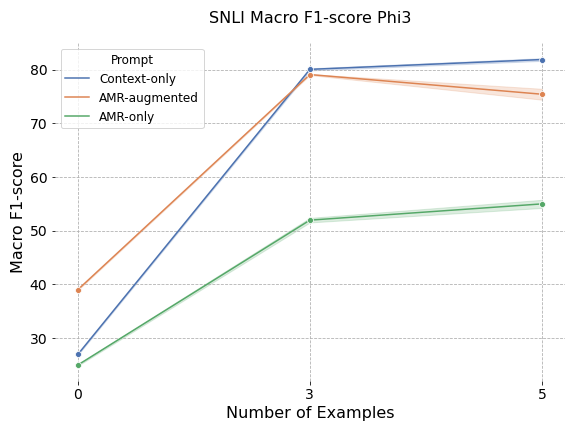}
    \caption{Phi3 SNLI Macro F1-score with 90\% confidence intervals.}
    \label{fig:snli_f1}
\end{figure}

Unlike the other tasks, for natural language inference on the SNLI dataset, Phi3 was the best performing model by a significant margin, achieving an 82\% macro F1-score with context-only 5-shot prompting. Fig. \ref{fig:snli_f1} shows that AMR-augmented prompting in Phi3 yields a significantly better zero-shot macro F1-score (39\%) compared to context-only prompting (27\%), which is only slightly better than AMR-only prompting (25\%). However, with the addition of few-shot examples in the prompt, context-only prompting achieves the highest macro F1-score. Detailed results for this task, including results from other models, are provided in Table \ref{tab:snli} in the Appendix.

\subsection{DocNLI}

\begin{figure}
    \centering
    \includegraphics[width=1\linewidth]{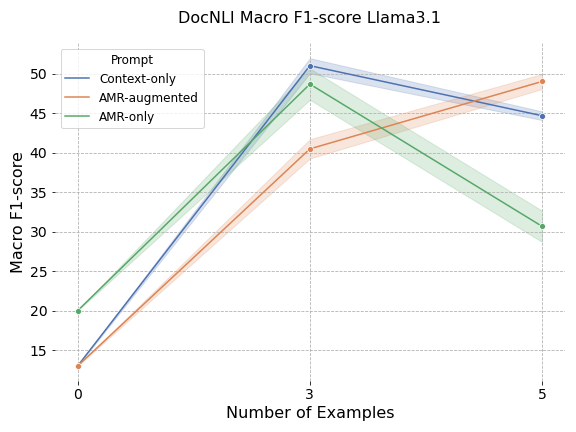}
    \caption{Llama3.1 DocNLI Macro F1-score with 90\% confidence intervals.}
    \label{fig:dnli_f1}
\end{figure}

The DocNLI experiment was conducted on approximately 13k examples from the DocNLI test split, which had an 8:5 label imbalance in favor of contradiction. This experiment aimed to further validate observations from prior experiments regarding the utility of AMRs in long-context tasks. Fig. \ref{fig:dnli_f1} illustrates that AMR-only prompting achieves the highest zero-shot macro F1 score of 20\%. However, with 3 few-shot examples, context-only prompting achieves a macro F1 score of 51\%, outperforming the other prompting methods. Both AMR-only and context-only prompting result in a performance decline with 5 few-shot examples, while the AMR-augmented prompting macro F1 score increases. These results are not only inconsistent with the SNLI experiment outcome but also deviate from the trends observed in other long-context tasks.

\section{Conclusion}

In this study, we investigated the ability of LLMs to interpret and leverage AMRs. Our findings demonstrate that AMR-augmented prompting, where the AMR of the context is included alongside the original context, significantly improves zero-shot long-dialogue summarization for Llama3.1 and noticeably improves sentence-level natural language inference for Phi3. The improvement from AMR-augmented prompting is more significant in SAMSum summarization with Llama3.1, the largest and most recently developed model among those used in this study, compared to other models and tasks. While AMRs did not enhance performance in other tasks, there was still some evidence that LLMs can extract meaningful information from AMRs. This was particularly evident in some of the AMR-only experiments, where the models achieved reasonable performance using AMRs alone. Additionally, we conducted rank-32 LoRA fine-tuning of Llama3.1 on the SAMSum summarization task, which produced results better than zero-shot but not as effective as few-shot prompting for both context-only and AMR-augmented approaches.

Overall, the experiments suggest that AMRs can assist LLMs in understanding long-term dependencies, key ideas, and events in long texts, as demonstrated by the zero-shot and few-shot SAMSum summarization experiments. However, including linearized AMRs in the prompt appears generally ineffective for tasks involving short contexts.

\section{Limitations}

In this work, we aimed to provide a comprehensive analysis of the impact of linearized AMRs on LLM performance. However, how these LLMs interpret AMRs after full fine-tuning on AMR-augmented and AMR-only tasks remains unclear, as it was outside the scope of this study. The only fine-tuning conducted in this study (LoRA fine-tuning of the 8-bit quantized Llama3.1) proved less effective than few-shot prompting. Moreover, the DocNLI long-context task was evaluated on only a partial test set. To establish confidence in the results for this task, the full test split of DocNLI should be evaluated. Another limitation of this work is that the HotpotQA experiments did not incorporate Chain-of-Thought prompting \cite{wei2023chainofthoughtpromptingelicitsreasoning}, which is the most commonly used prompting technique for achieving strong performance on this dataset. This omission is justified, however, given that Chain-of-Thought prompting tends to work well only with models larger than those used in this study.

Our results demonstrate that AMRs improve performance on long-context tasks such as dialogue summarization but degrade performance on short-context tasks like single-hop QA. However, we have not conducted a detailed analysis to identify the underlying causes. Further investigation is needed to uncover task-specific factors that contribute to these outcomes.

\section{Future Work}

There are numerous directions for extending this work. For instance, approaches such as prompt tuning through soft prompts \cite{lester2021powerscaleparameterefficientprompt}, synthetic AMR generation and AMR enrichment \cite{ji-etal-2022-automatic}, or adapter-based parameter-efficient fine-tuning \cite{houlsby2019parameterefficienttransferlearningnlp} can be explored alongside full fine-tuning of LLMs on AMR-augmented and AMR-only tasks. This study utilized only one of the most recently developed LLMs; analyzing other newer models as large as Llama3.1 or larger could help reinforce the observations presented in this paper. Additionally, exploring how AMRs or other structured representations impact retrieval augmented pipelines would be an interesting avenue for future research. It would also be interesting to observe how the findings of this paper change when the models are 4-bit quantized.

The analysis presented in this work can also be extended to other structured semantic representations using similar tasks. In particular, evaluating Discourse Representation Structures (DRS) and Knowledge Graphs (KG) would be valuable, given the wide range of tasks for which these representations are continuously being explored.

\bibliography{custom}

\begin{thebibliography}{22}
\providecommand{\natexlab}[1]{#1}

\bibitem[{Abdin et~al.(2024)Abdin, Aneja, Awadalla, Awadallah, Awan, Bach, Bahree, Bakhtiari, Bao, Behl, Benhaim, Bilenko, Bjorck, Bubeck, Cai, Cai, Chaudhary, Chen, Chen, Chen, Chen, Chen, Cheng, Chopra, Dai, Dixon, Eldan, Fragoso, Gao, Gao, Gao, Garg, Giorno, Goswami, Gunasekar, Haider, Hao, Hewett, Hu, Huynh, Iter, Jacobs, Javaheripi, Jin, Karampatziakis, Kauffmann, Khademi, Kim, Kim, Kurilenko, Lee, Lee, Li, Li, Liang, Liden, Lin, Lin, Liu, Liu, Liu, Liu, Liu, Luo, Madan, Mahmoudzadeh, Majercak, Mazzola, Mendes, Mitra, Modi, Nguyen, Norick, Patra, Perez-Becker, Portet, Pryzant, Qin, Radmilac, Ren, de~Rosa, Rosset, Roy, Ruwase, Saarikivi, Saied, Salim, Santacroce, Shah, Shang, Sharma, Shen, Shukla, Song, Tanaka, Tupini, Vaddamanu, Wang, Wang, Wang, Wang, Wang, Wang, Ward, Wen, Witte, Wu, Wu, Wyatt, Xiao, Xu, Xu, Xu, Xue, Yadav, Yang, Yang, Yang, Yang, Yu, Yuan, Zhang, Zhang, Zhang, Zhang, Zhang, Zhang, Zhang, and Zhou}]{abdin2024phi3technicalreporthighly}
Marah Abdin, Jyoti Aneja, Hany Awadalla, Ahmed Awadallah, Ammar~Ahmad Awan, Nguyen Bach, Amit Bahree, Arash Bakhtiari, Jianmin Bao, Harkirat Behl, Alon Benhaim, Misha Bilenko, Johan Bjorck, Sébastien Bubeck, Martin Cai, Qin Cai, Vishrav Chaudhary, Dong Chen, Dongdong Chen, and 110 others. 2024.
\newblock \href {https://arxiv.org/abs/2404.14219} {Phi-3 technical report: A highly capable language model locally on your phone}.
\newblock \emph{Preprint}, arXiv:2404.14219.

\bibitem[{Bowman et~al.(2015)Bowman, Angeli, Potts, and Manning}]{bowman2015largeannotatedcorpuslearning}
Samuel~R. Bowman, Gabor Angeli, Christopher Potts, and Christopher~D. Manning. 2015.
\newblock \href {https://arxiv.org/abs/1508.05326} {A large annotated corpus for learning natural language inference}.
\newblock \emph{Preprint}, arXiv:1508.05326.

\bibitem[{Dong and Holder(2014)}]{Dong2014NaturalLG}
Ngan~T. Dong and Lawrence~B. Holder. 2014.
\newblock \href {https://api.semanticscholar.org/CorpusID:31393024} {Natural language generation from graphs}.
\newblock \emph{Int. J. Semantic Comput.}, 8:335--.

\bibitem[{Drozdov et~al.(2022)Drozdov, Zhou, Florian, McCallum, Naseem, Kim, and Astudillo}]{drozdov2022inducingusingalignmentstransitionbased}
Andrew Drozdov, Jiawei Zhou, Radu Florian, Andrew McCallum, Tahira Naseem, Yoon Kim, and Ramon~Fernandez Astudillo. 2022.
\newblock \href {https://arxiv.org/abs/2205.01464} {Inducing and using alignments for transition-based amr parsing}.
\newblock \emph{Preprint}, arXiv:2205.01464.

\bibitem[{Gliwa et~al.(2019)Gliwa, Mochol, Biesek, and Wawer}]{gliwa-etal-2019-samsum}
Bogdan Gliwa, Iwona Mochol, Maciej Biesek, and Aleksander Wawer. 2019.
\newblock \href {https://doi.org/10.18653/v1/D19-5409} {{SAMS}um corpus: A human-annotated dialogue dataset for abstractive summarization}.
\newblock In \emph{Proceedings of the 2nd Workshop on New Frontiers in Summarization}, pages 70--79, Hong Kong, China. Association for Computational Linguistics.

\bibitem[{Grattafiori et~al.(2024)Grattafiori, Dubey, Jauhri, Pandey, Kadian, Al-Dahle, Letman, Mathur, Schelten, Vaughan, Yang, Fan, Goyal, Hartshorn, Yang, Mitra, Sravankumar, Korenev, Hinsvark, Rao, Zhang, Rodriguez, Gregerson, Spataru, Roziere, Biron, Tang, Chern, Caucheteux, Nayak, Bi, Marra, McConnell, Keller, Touret, Wu, Wong, Ferrer, Nikolaidis, Allonsius, Song, Pintz, Livshits, Wyatt, Esiobu, Choudhary, Mahajan, Garcia-Olano, Perino, Hupkes, Lakomkin, AlBadawy, Lobanova, Dinan, Smith, Radenovic, Guzmán, Zhang, Synnaeve, Lee, Anderson, Thattai, Nail, Mialon, Pang, Cucurell, Nguyen, Korevaar, Xu, Touvron, Zarov, Ibarra, Kloumann, Misra, Evtimov, Zhang, Copet, Lee, Geffert, Vranes, Park, Mahadeokar, Shah, van~der Linde, Billock, Hong, Lee, Fu, Chi, Huang, Liu, Wang, Yu, Bitton, Spisak, Park, Rocca, Johnstun, Saxe, Jia, Alwala, Prasad, Upasani, Plawiak, Li, Heafield, Stone, El-Arini, Iyer, Malik, Chiu, Bhalla, Lakhotia, Rantala-Yeary, van~der Maaten, Chen, Tan, Jenkins, Martin, Madaan, Malo, Blecher,
  Landzaat, de~Oliveira, Muzzi, Pasupuleti, Singh, Paluri, Kardas, Tsimpoukelli, Oldham, Rita, Pavlova, Kambadur, Lewis, Si, Singh, Hassan, Goyal, Torabi, Bashlykov, Bogoychev, Chatterji, Zhang, Duchenne, Çelebi, Alrassy, Zhang, Li, Vasic, Weng, Bhargava, Dubal, Krishnan, Koura, Xu, He, Dong, Srinivasan, Ganapathy, Calderer, Cabral, Stojnic, Raileanu, Maheswari, Girdhar, Patel, Sauvestre, Polidoro, Sumbaly, Taylor, Silva, Hou, Wang, Hosseini, Chennabasappa, Singh, Bell, Kim, Edunov, Nie, Narang, Raparthy, Shen, Wan, Bhosale, Zhang, Vandenhende, Batra, Whitman, Sootla, Collot, Gururangan, Borodinsky, Herman, Fowler, Sheasha, Georgiou, Scialom, Speckbacher, Mihaylov, Xiao, Karn, Goswami, Gupta, Ramanathan, Kerkez, Gonguet, Do, Vogeti, Albiero, Petrovic, Chu, Xiong, Fu, Meers, Martinet, Wang, Wang, Tan, Xia, Xie, Jia, Wang, Goldschlag, Gaur, Babaei, Wen, Song, Zhang, Li, Mao, Coudert, Yan, Chen, Papakipos, Singh, Srivastava, Jain, Kelsey, Shajnfeld, Gangidi, Victoria, Goldstand, Menon, Sharma, Boesenberg,
  Baevski, Feinstein, Kallet, Sangani, Teo, Yunus, Lupu, Alvarado, Caples, Gu, Ho, Poulton, Ryan, Ramchandani, Dong, Franco, Goyal, Saraf, Chowdhury, Gabriel, Bharambe, Eisenman, Yazdan, James, Maurer, Leonhardi, Huang, Loyd, Paola, Paranjape, Liu, Wu, Ni, Hancock, Wasti, Spence, Stojkovic, Gamido, Montalvo, Parker, Burton, Mejia, Liu, Wang, Kim, Zhou, Hu, Chu, Cai, Tindal, Feichtenhofer, Gao, Civin, Beaty, Kreymer, Li, Adkins, Xu, Testuggine, David, Parikh, Liskovich, Foss, Wang, Le, Holland, Dowling, Jamil, Montgomery, Presani, Hahn, Wood, Le, Brinkman, Arcaute, Dunbar, Smothers, Sun, Kreuk, Tian, Kokkinos, Ozgenel, Caggioni, Kanayet, Seide, Florez, Schwarz, Badeer, Swee, Halpern, Herman, Sizov, Guangyi, Zhang, Lakshminarayanan, Inan, Shojanazeri, Zou, Wang, Zha, Habeeb, Rudolph, Suk, Aspegren, Goldman, Zhan, Damlaj, Molybog, Tufanov, Leontiadis, Veliche, Gat, Weissman, Geboski, Kohli, Lam, Asher, Gaya, Marcus, Tang, Chan, Zhen, Reizenstein, Teboul, Zhong, Jin, Yang, Cummings, Carvill, Shepard, McPhie,
  Torres, Ginsburg, Wang, Wu, U, Saxena, Khandelwal, Zand, Matosich, Veeraraghavan, Michelena, Li, Jagadeesh, Huang, Chawla, Huang, Chen, Garg, A, Silva, Bell, Zhang, Guo, Yu, Moshkovich, Wehrstedt, Khabsa, Avalani, Bhatt, Mankus, Hasson, Lennie, Reso, Groshev, Naumov, Lathi, Keneally, Liu, Seltzer, Valko, Restrepo, Patel, Vyatskov, Samvelyan, Clark, Macey, Wang, Hermoso, Metanat, Rastegari, Bansal, Santhanam, Parks, White, Bawa, Singhal, Egebo, Usunier, Mehta, Laptev, Dong, Cheng, Chernoguz, Hart, Salpekar, Kalinli, Kent, Parekh, Saab, Balaji, Rittner, Bontrager, Roux, Dollar, Zvyagina, Ratanchandani, Yuvraj, Liang, Alao, Rodriguez, Ayub, Murthy, Nayani, Mitra, Parthasarathy, Li, Hogan, Battey, Wang, Howes, Rinott, Mehta, Siby, Bondu, Datta, Chugh, Hunt, Dhillon, Sidorov, Pan, Mahajan, Verma, Yamamoto, Ramaswamy, Lindsay, Lindsay, Feng, Lin, Zha, Patil, Shankar, Zhang, Zhang, Wang, Agarwal, Sajuyigbe, Chintala, Max, Chen, Kehoe, Satterfield, Govindaprasad, Gupta, Deng, Cho, Virk, Subramanian, Choudhury,
  Goldman, Remez, Glaser, Best, Koehler, Robinson, Li, Zhang, Matthews, Chou, Shaked, Vontimitta, Ajayi, Montanez, Mohan, Kumar, Mangla, Ionescu, Poenaru, Mihailescu, Ivanov, Li, Wang, Jiang, Bouaziz, Constable, Tang, Wu, Wang, Wu, Gao, Kleinman, Chen, Hu, Jia, Qi, Li, Zhang, Zhang, Adi, Nam, Yu, Wang, Zhao, Hao, Qian, Li, He, Rait, DeVito, Rosnbrick, Wen, Yang, Zhao, and Ma}]{grattafiori2024llama3herdmodels}
Aaron Grattafiori, Abhimanyu Dubey, Abhinav Jauhri, Abhinav Pandey, Abhishek Kadian, Ahmad Al-Dahle, Aiesha Letman, Akhil Mathur, Alan Schelten, Alex Vaughan, Amy Yang, Angela Fan, Anirudh Goyal, Anthony Hartshorn, Aobo Yang, Archi Mitra, Archie Sravankumar, Artem Korenev, Arthur Hinsvark, and 542 others. 2024.
\newblock \href {https://arxiv.org/abs/2407.21783} {The llama 3 herd of models}.
\newblock \emph{Preprint}, arXiv:2407.21783.

\bibitem[{Houlsby et~al.(2019)Houlsby, Giurgiu, Jastrzebski, Morrone, de~Laroussilhe, Gesmundo, Attariyan, and Gelly}]{houlsby2019parameterefficienttransferlearningnlp}
Neil Houlsby, Andrei Giurgiu, Stanislaw Jastrzebski, Bruna Morrone, Quentin de~Laroussilhe, Andrea Gesmundo, Mona Attariyan, and Sylvain Gelly. 2019.
\newblock \href {https://arxiv.org/abs/1902.00751} {Parameter-efficient transfer learning for nlp}.
\newblock \emph{Preprint}, arXiv:1902.00751.

\bibitem[{Hu et~al.(2021)Hu, Shen, Wallis, Allen-Zhu, Li, Wang, Wang, and Chen}]{hu2021loralowrankadaptationlarge}
Edward~J. Hu, Yelong Shen, Phillip Wallis, Zeyuan Allen-Zhu, Yuanzhi Li, Shean Wang, Lu~Wang, and Weizhu Chen. 2021.
\newblock \href {https://arxiv.org/abs/2106.09685} {Lora: Low-rank adaptation of large language models}.
\newblock \emph{Preprint}, arXiv:2106.09685.

\bibitem[{Hua et~al.(2023)Hua, Deng, and McKeown}]{hua-etal-2023-improving}
Yilun Hua, Zhaoyuan Deng, and Kathleen McKeown. 2023.
\newblock \href {https://doi.org/10.18653/v1/2023.findings-acl.871} {Improving long dialogue summarization with semantic graph representation}.
\newblock In \emph{Findings of the Association for Computational Linguistics: ACL 2023}, pages 13851--13883, Toronto, Canada. Association for Computational Linguistics.

\bibitem[{Ji et~al.(2022)Ji, Williamson, and Choi}]{ji-etal-2022-automatic}
Yuxin Ji, Gregor Williamson, and Jinho~D. Choi. 2022.
\newblock \href {https://aclanthology.org/2022.law-1.19/} {Automatic enrichment of {A}bstract {M}eaning {R}epresentations}.
\newblock In \emph{Proceedings of the 16th Linguistic Annotation Workshop (LAW-XVI) within LREC2022}, pages 160--169, Marseille, France. European Language Resources Association.

\bibitem[{Jiang et~al.(2023)Jiang, Sablayrolles, Mensch, Bamford, Chaplot, de~las Casas, Bressand, Lengyel, Lample, Saulnier, Lavaud, Lachaux, Stock, Scao, Lavril, Wang, Lacroix, and Sayed}]{jiang2023mistral7b}
Albert~Q. Jiang, Alexandre Sablayrolles, Arthur Mensch, Chris Bamford, Devendra~Singh Chaplot, Diego de~las Casas, Florian Bressand, Gianna Lengyel, Guillaume Lample, Lucile Saulnier, Lélio~Renard Lavaud, Marie-Anne Lachaux, Pierre Stock, Teven~Le Scao, Thibaut Lavril, Thomas Wang, Timothée Lacroix, and William~El Sayed. 2023.
\newblock \href {https://arxiv.org/abs/2310.06825} {Mistral 7b}.
\newblock \emph{Preprint}, arXiv:2310.06825.

\bibitem[{Ke et~al.(2021)Ke, Ji, Ran, Cui, Wang, Song, Zhu, and Huang}]{ke2021jointgtgraphtextjointrepresentation}
Pei Ke, Haozhe Ji, Yu~Ran, Xin Cui, Liwei Wang, Linfeng Song, Xiaoyan Zhu, and Minlie Huang. 2021.
\newblock \href {https://arxiv.org/abs/2106.10502} {Jointgt: Graph-text joint representation learning for text generation from knowledge graphs}.
\newblock \emph{Preprint}, arXiv:2106.10502.

\bibitem[{Koncel-Kedziorski et~al.(2022)Koncel-Kedziorski, Bekal, Luan, Lapata, and Hajishirzi}]{koncelkedziorski2022textgenerationknowledgegraphs}
Rik Koncel-Kedziorski, Dhanush Bekal, Yi~Luan, Mirella Lapata, and Hannaneh Hajishirzi. 2022.
\newblock \href {https://arxiv.org/abs/1904.02342} {Text generation from knowledge graphs with graph transformers}.
\newblock \emph{Preprint}, arXiv:1904.02342.

\bibitem[{Langkilde and Knight(1998)}]{langkilde-knight-1998-generation-exploits}
Irene Langkilde and Kevin Knight. 1998.
\newblock \href {https://doi.org/10.3115/980845.980963} {Generation that exploits corpus-based statistical knowledge}.
\newblock In \emph{36th Annual Meeting of the Association for Computational Linguistics and 17th International Conference on Computational Linguistics, Volume 1}, pages 704--710, Montreal, Quebec, Canada. Association for Computational Linguistics.

\bibitem[{Lester et~al.(2021)Lester, Al-Rfou, and Constant}]{lester2021powerscaleparameterefficientprompt}
Brian Lester, Rami Al-Rfou, and Noah Constant. 2021.
\newblock \href {https://arxiv.org/abs/2104.08691} {The power of scale for parameter-efficient prompt tuning}.
\newblock \emph{Preprint}, arXiv:2104.08691.

\bibitem[{Rajpurkar et~al.(2018)Rajpurkar, Jia, and Liang}]{rajpurkar-etal-2018-know}
Pranav Rajpurkar, Robin Jia, and Percy Liang. 2018.
\newblock \href {https://doi.org/10.18653/v1/P18-2124} {Know what you don`t know: Unanswerable questions for {SQ}u{AD}}.
\newblock In \emph{Proceedings of the 56th Annual Meeting of the Association for Computational Linguistics (Volume 2: Short Papers)}, pages 784--789, Melbourne, Australia. Association for Computational Linguistics.

\bibitem[{Wang et~al.(2020)Wang, Wei, Dong, Bao, Yang, and Zhou}]{wang2020minilmdeepselfattentiondistillation}
Wenhui Wang, Furu Wei, Li~Dong, Hangbo Bao, Nan Yang, and Ming Zhou. 2020.
\newblock \href {https://arxiv.org/abs/2002.10957} {Minilm: Deep self-attention distillation for task-agnostic compression of pre-trained transformers}.
\newblock \emph{Preprint}, arXiv:2002.10957.

\bibitem[{Wei et~al.(2023)Wei, Wang, Schuurmans, Bosma, Ichter, Xia, Chi, Le, and Zhou}]{wei2023chainofthoughtpromptingelicitsreasoning}
Jason Wei, Xuezhi Wang, Dale Schuurmans, Maarten Bosma, Brian Ichter, Fei Xia, Ed~Chi, Quoc Le, and Denny Zhou. 2023.
\newblock \href {https://arxiv.org/abs/2201.11903} {Chain-of-thought prompting elicits reasoning in large language models}.
\newblock \emph{Preprint}, arXiv:2201.11903.

\bibitem[{Yang et~al.(2024)Yang, Zhao, Tang, Liu, Zhan, and Lin}]{yang2024emphasisingstructuredinformationintegrating}
Bohao Yang, Kun Zhao, Chen Tang, Dong Liu, Liang Zhan, and Chenghua Lin. 2024.
\newblock \href {https://arxiv.org/abs/2404.01129} {Emphasising structured information: Integrating abstract meaning representation into llms for enhanced open-domain dialogue evaluation}.
\newblock \emph{Preprint}, arXiv:2404.01129.

\bibitem[{Yang et~al.(2018)Yang, Qi, Zhang, Bengio, Cohen, Salakhutdinov, and Manning}]{yang2018hotpotqadatasetdiverseexplainable}
Zhilin Yang, Peng Qi, Saizheng Zhang, Yoshua Bengio, William~W. Cohen, Ruslan Salakhutdinov, and Christopher~D. Manning. 2018.
\newblock \href {https://arxiv.org/abs/1809.09600} {Hotpotqa: A dataset for diverse, explainable multi-hop question answering}.
\newblock \emph{Preprint}, arXiv:1809.09600.

\bibitem[{Yin et~al.(2021)Yin, Radev, and Xiong}]{yin-etal-2021-docnli}
Wenpeng Yin, Dragomir Radev, and Caiming Xiong. 2021.
\newblock \href {https://doi.org/10.18653/v1/2021.findings-acl.435} {{D}oc{NLI}: A large-scale dataset for document-level natural language inference}.
\newblock In \emph{Findings of the Association for Computational Linguistics: ACL-IJCNLP 2021}, pages 4913--4922, Online. Association for Computational Linguistics.

\bibitem[{Zhu et~al.(2019)Zhu, Li, Zhu, Qian, Zhang, and Zhou}]{zhu-etal-2019-modeling}
Jie Zhu, Junhui Li, Muhua Zhu, Longhua Qian, Min Zhang, and Guodong Zhou. 2019.
\newblock \href {https://doi.org/10.18653/v1/D19-1548} {Modeling graph structure in transformer for better {AMR}-to-text generation}.
\newblock In \emph{Proceedings of the 2019 Conference on Empirical Methods in Natural Language Processing and the 9th International Joint Conference on Natural Language Processing (EMNLP-IJCNLP)}, pages 5459--5468, Hong Kong, China. Association for Computational Linguistics.

\end{thebibliography}

\appendix
\section{Appendix}

\begin{figure}
    \centering
    \includegraphics[width=1\linewidth]{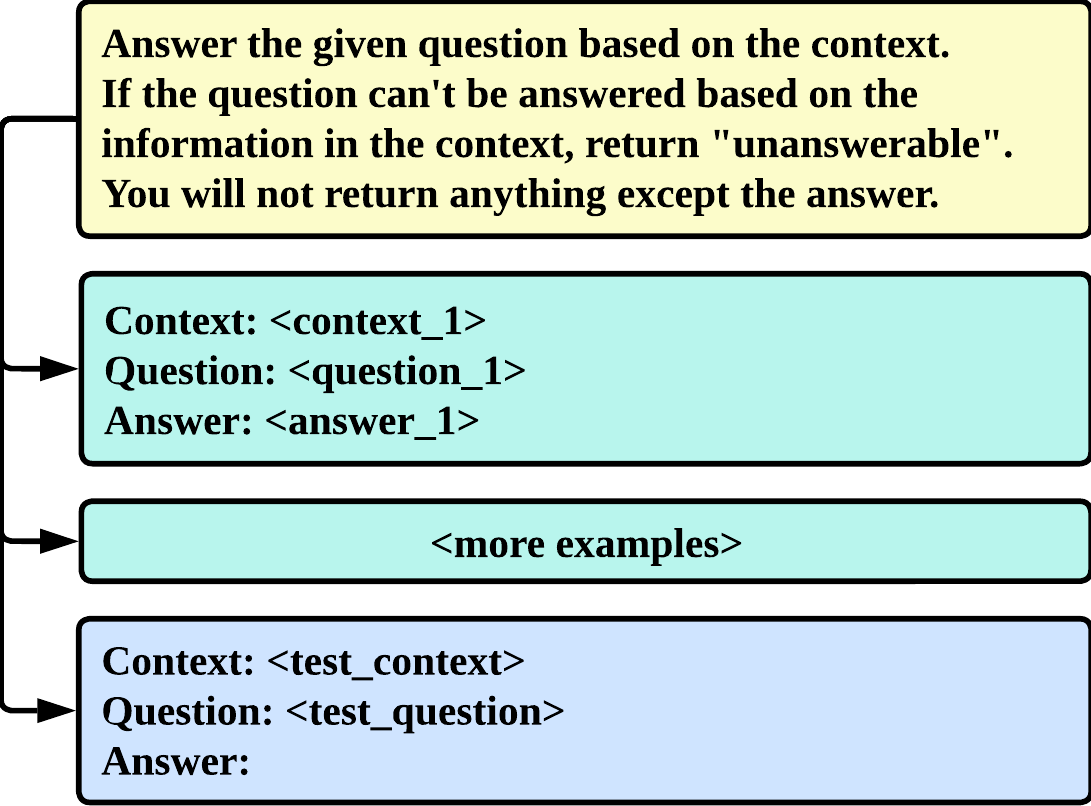}
    \caption{Context-only prompt for QA on the SQuAD 2.0 dataset.}
    \label{fig:squad_raw_prompt}
\end{figure}

\begin{figure}
    \centering
    \includegraphics[width=1\linewidth]{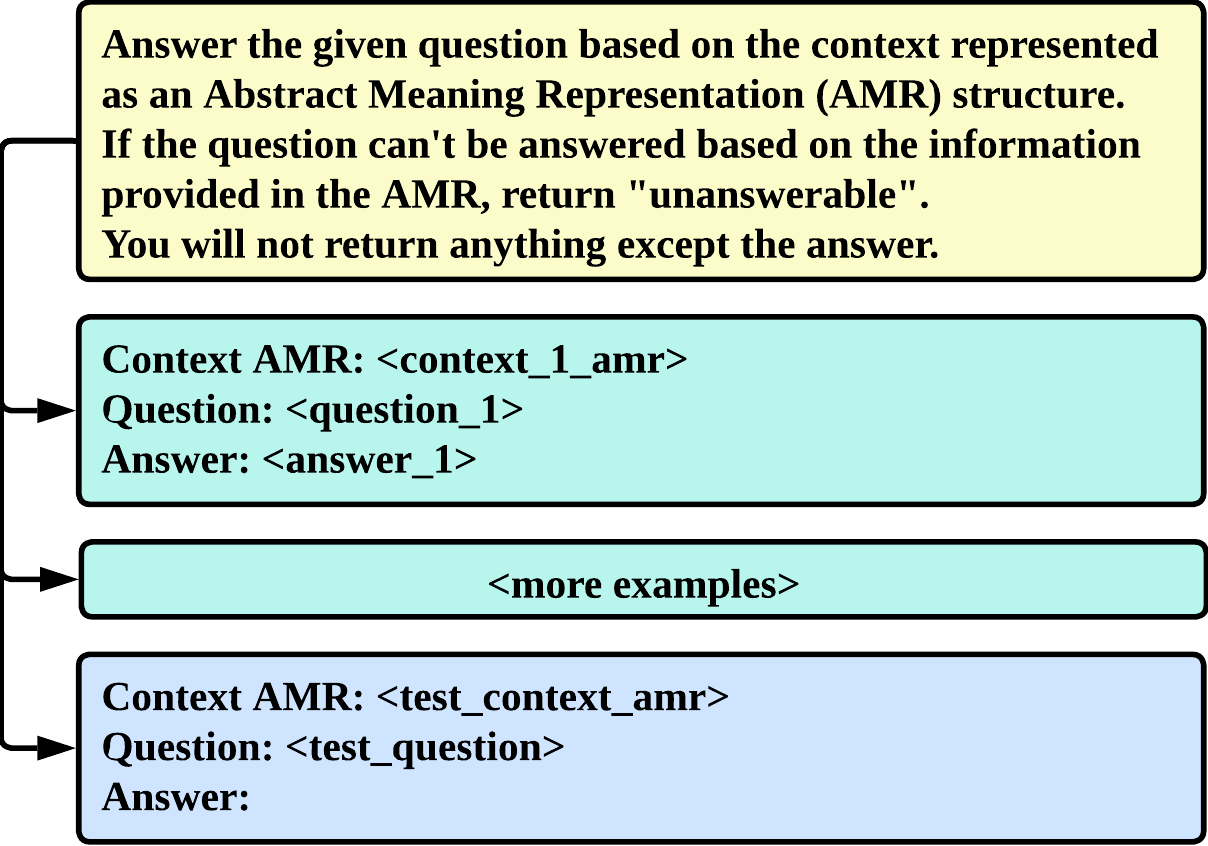}
    \caption{AMR-only prompt for QA on SQuAD 2.0 dataset.}
    \label{fig:squad_oamr_prompt}
\end{figure}

\begin{figure}
    \centering
    \includegraphics[width=1\linewidth]{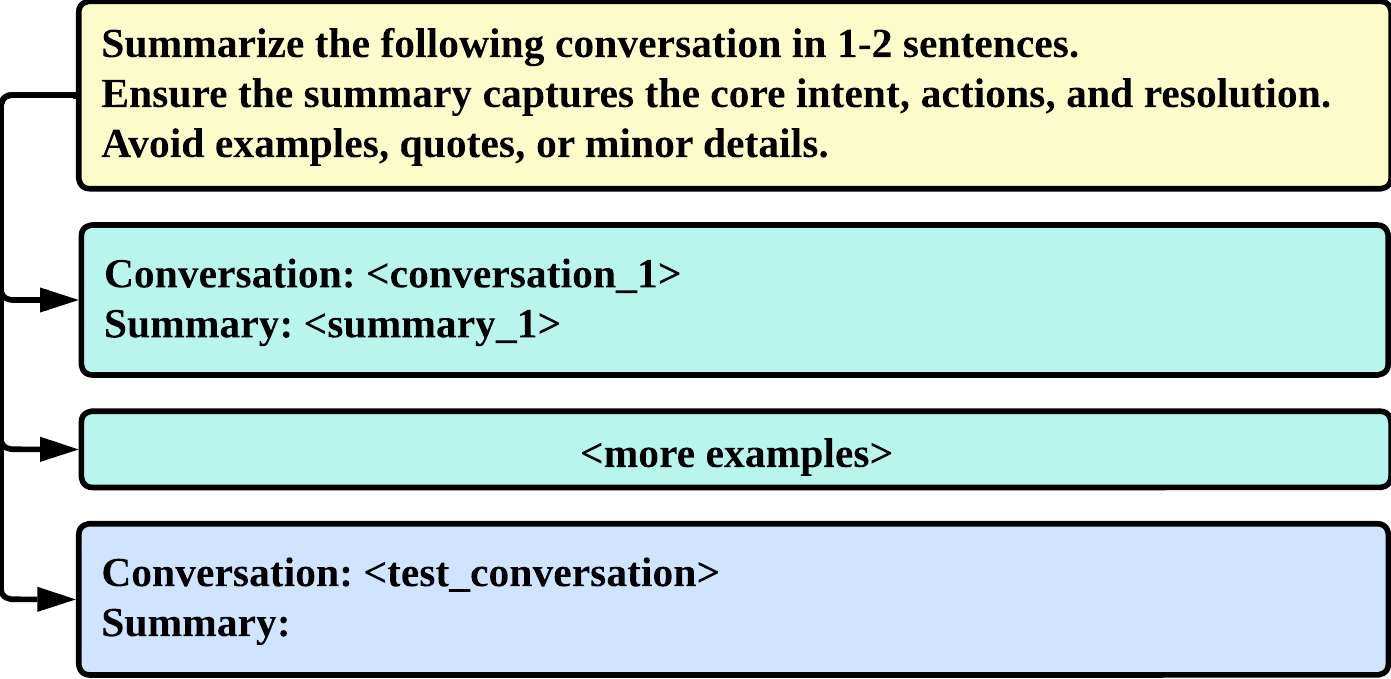}
    \caption{Context-only prompt for summarizing SAMSum dataset conversations.}
    \label{fig:samsum_raw_prompt}
\end{figure}

\begin{figure}
    \centering
    \includegraphics[width=1\linewidth]{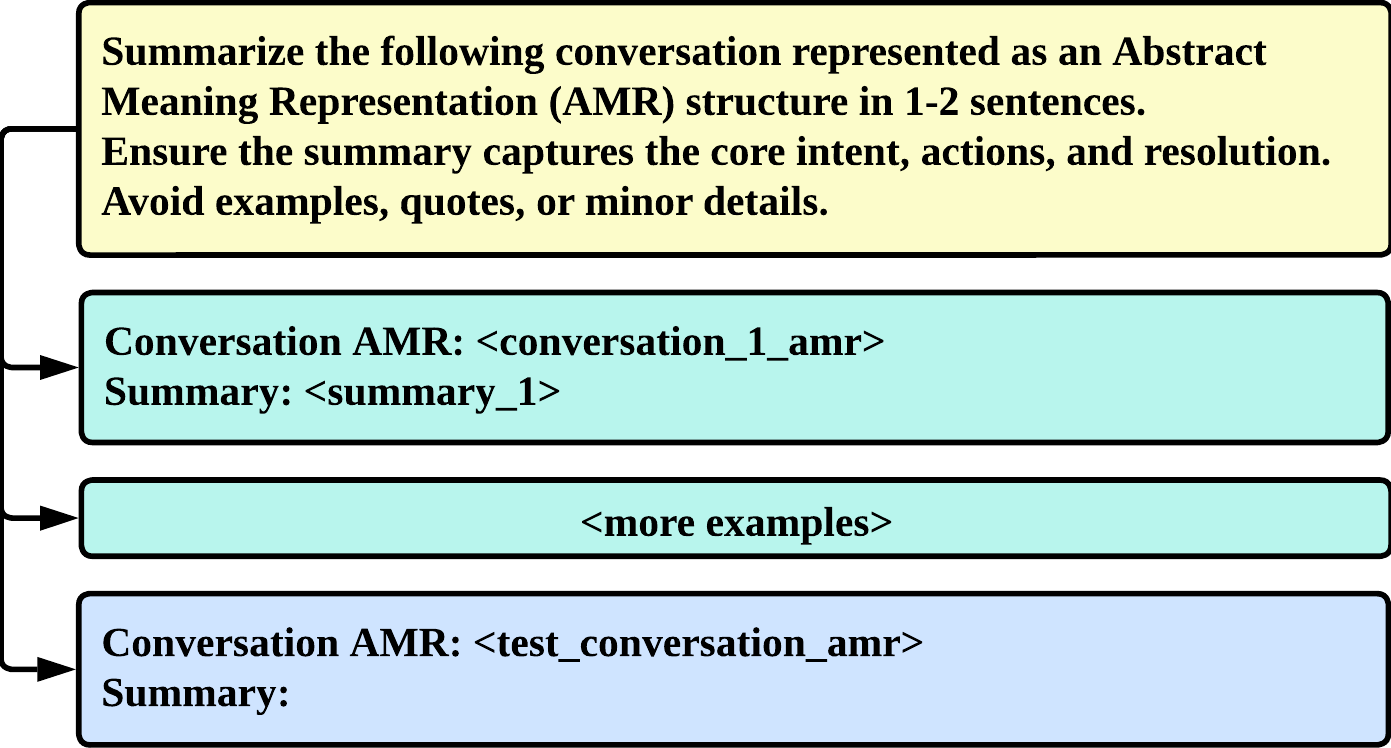}
    \caption{AMR-only prompt for summarizing SAMSum dataset conversations.}
    \label{fig:samsum_oamr_prompt}
\end{figure}

\begin{figure}
    \centering
    \includegraphics[width=1\linewidth]{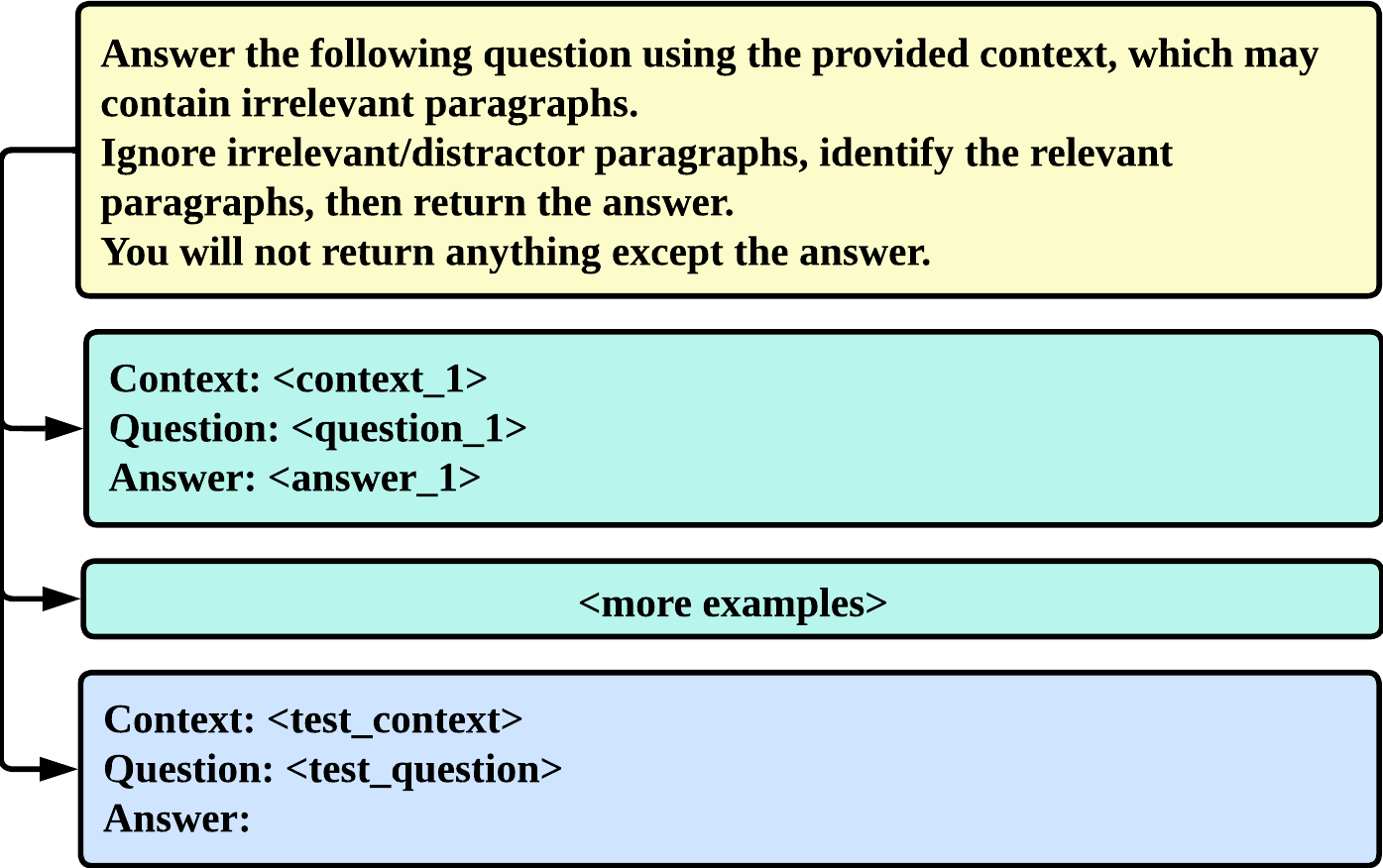}
    \caption{Context-only prompt for QA on HotpotQA dataset.}
    \label{fig:hotpot_raw_prompt}
\end{figure}

\begin{figure}
    \centering
    \includegraphics[width=1\linewidth]{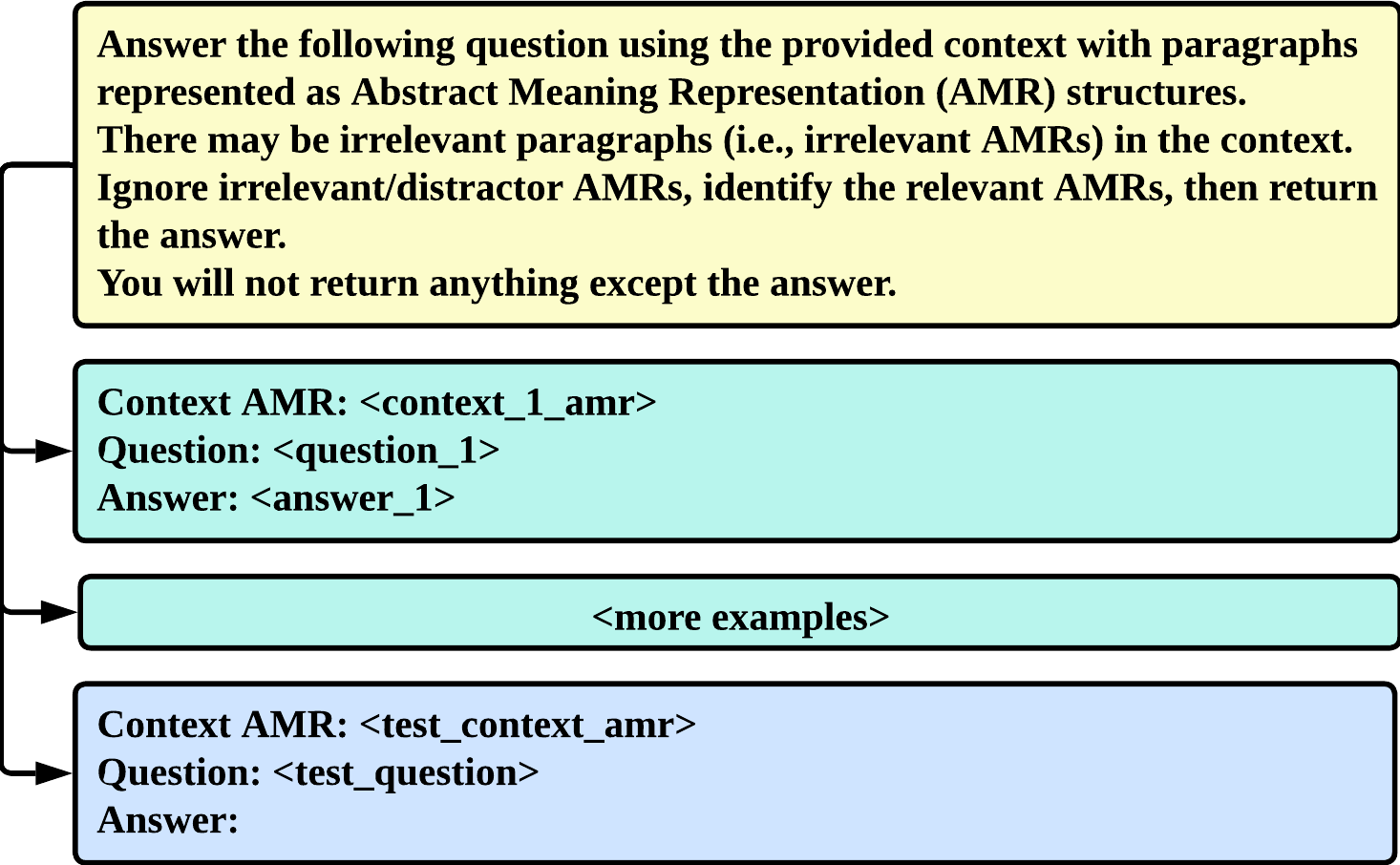}
    \caption{AMR-only prompt for QA on HotpotQA dataset.}
    \label{fig:hotpot_oamr_prompt}
\end{figure}

\begin{figure}
    \centering
    \includegraphics[width=1\linewidth]{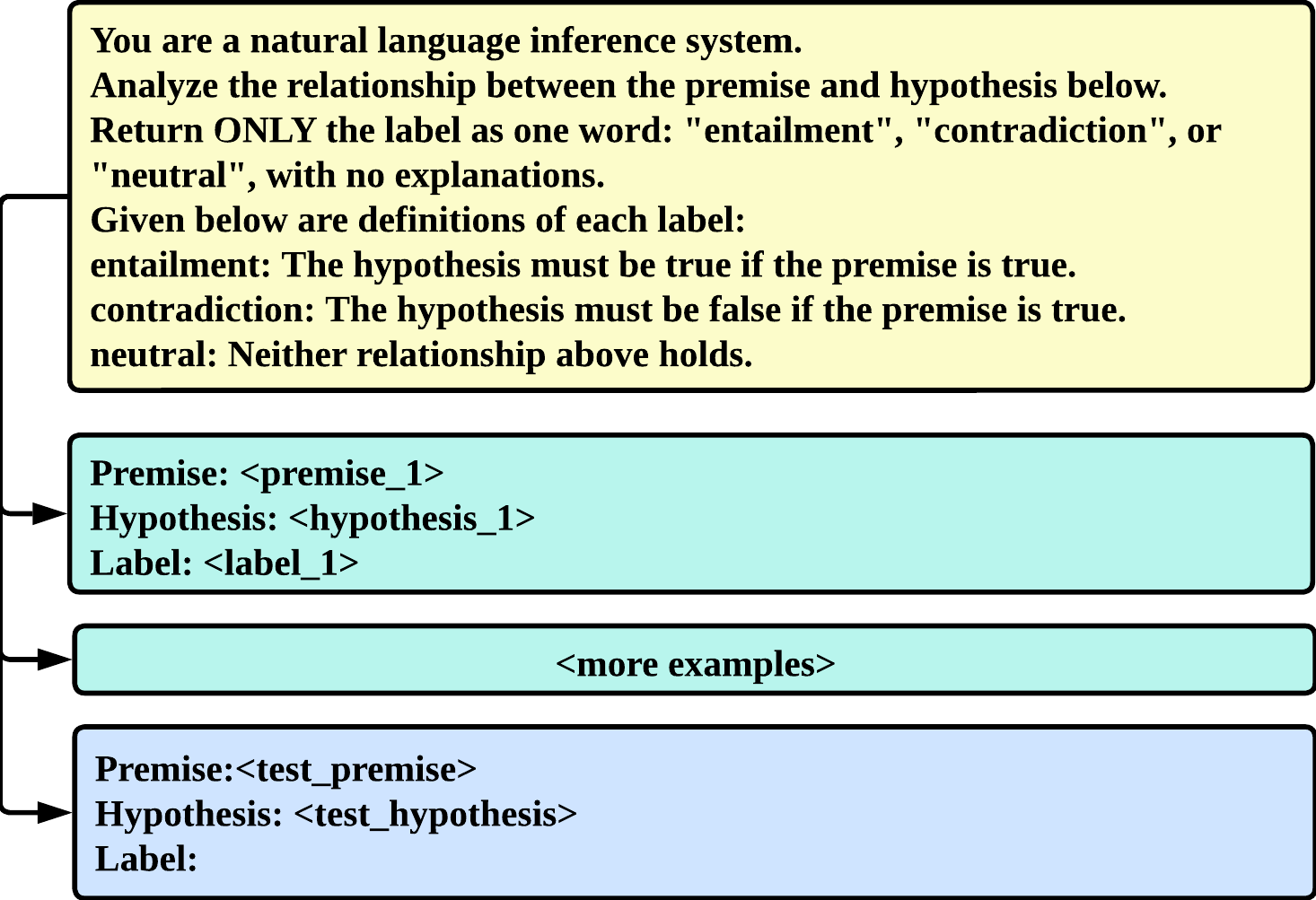}
    \caption{Context-only prompt for natural language inference on SNLI dataset.}
    \label{fig:snli_raw_prompt}
\end{figure}

\begin{figure}
    \centering
    \includegraphics[width=1\linewidth]{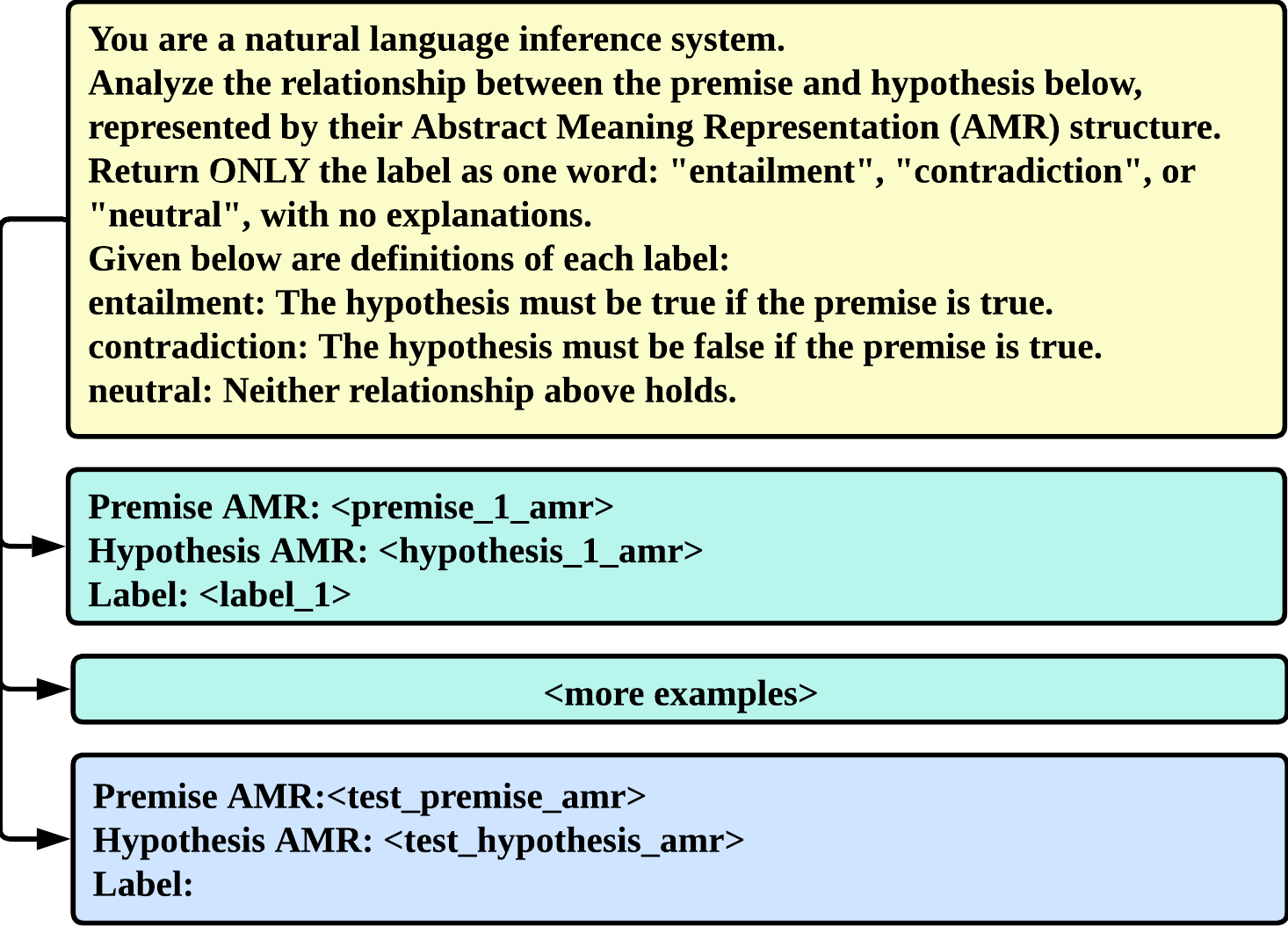}
    \caption{AMR-only prompt for natural language inference on SNLI dataset.}
    \label{fig:snli_oamr_prompt}
\end{figure}

\begin{figure}
    \centering
    \includegraphics[width=1\linewidth]{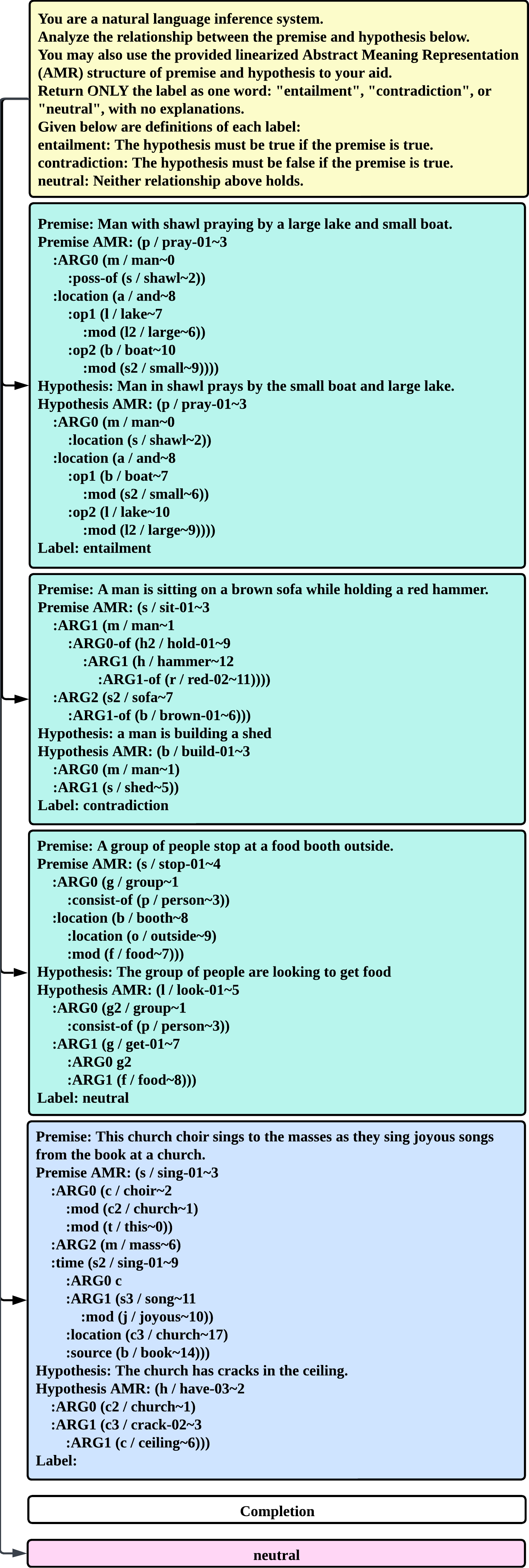}
    \caption{An example of a prompt-completion for the AMR-augmented 3-shot SNLI entailment prediction task using Llama3.1.}
    \label{fig:full_ex}
\end{figure}

\begin{table*}
\centering
\small
\caption{LDC2020T02 AMR-to-text full results.}
\label{tab:ldc}
\begin{tabular}{|l| c| c| c| c |c| c|}
\hline
\textbf{Model} & \textbf{\# Examples} & \textbf{Cosine similarity} & \textbf{ROUGE-1} & \textbf{ROUGE-2} & \textbf{ROUGE-L} & \textbf{BLEU} \\
\hline
{\textbf{Llama3.1}} & 0 & 73 & 55 & 20 & 43 & 6 \\
                          & 3 & 80 & 63 & 28 & 51 & 11 \\
                          & 5 & \textbf{81} & \textbf{64} & \textbf{30} & \textbf{52} & \textbf{12} \\
\hline
{\textbf{Phi3}}     & 0 & 74 & 55 & 21 & 43 & 6 \\
                          & 3 & 75 & 56 & 22 & 45 & 7 \\
                          & 5 & 76 & 57 & 23 & 45 & 8 \\
\hline
{\textbf{Mistral}}  & 0 & 69 & 51 & 18 & 40 & 5 \\
                          & 3 & 77 & 59 & 23 & 46 & 7 \\
                          & 5 & 77 & 59 & 25 & 47 & 9 \\
\hline
\end{tabular}
\end{table*}

\begin{table*}
\centering
\small
\caption{SQuAD 2.0 QA full results. 5-shot prompting was only conducted for the best performing model (Llama3.1).}
\label{tab:squad}
\begin{tabular}{|l| c| l| c| c|}
\hline
\textbf{Model} & \textbf{\# Examples} & \textbf{Prompt} & \textbf{F1} & \textbf{Cosine similarity} \\
\hline
{\textbf{Llama3.1}} & 0 & Context-only & 55 & 66 \\
                          & 3 & Context-only & 59 & 68 \\
                          & 5 & Context-only & \textbf{59} & \textbf{68} \\
\cline{2-5}
                          & 0 & AMR-augmented & 49 & 63 \\
                          & 3 & AMR-augmented & 52 & 62 \\
                          & 5 & AMR-augmented & 49 & 60 \\
\cline{2-5}
                          & 0 & AMR-only      & 18 & 38 \\
                          & 3 & AMR-only      & 48 & 60 \\
                          & 5 & AMR-only      & 26 & 45 \\
\hline
{\textbf{Phi3}}     & 0 & Context-only & 46 & 56 \\
                          & 3 & Context-only & \textbf{47} & \textbf{58} \\
\cline{2-5}
                          & 0 & AMR-augmented & 38 & 46 \\
                          & 3 & AMR-augmented & 40 & 53 \\
\cline{2-5}
                          & 0 & AMR-only      & 20 & 40 \\
                          & 3 & AMR-only      & 22 & 42 \\
\hline
{\textbf{Mistral}}  & 0 & Context-only & 35 & 47 \\
                          & 3 & Context-only & \textbf{51} & \textbf{61} \\
\cline{2-5}
                          & 0 & AMR-augmented & 27 & 37 \\
                          & 3 & AMR-augmented & 49 & 60 \\
\cline{2-5}
                          & 0 & AMR-only      & 17 & 35 \\
                          & 3 & AMR-only      & 41 & 55 \\
\hline
\end{tabular}
\end{table*}

\begin{table*}
\centering
\small
\caption{SAMSum summarization full results. 5-shot prompting and LoRA fine-tuning were only conducted for the model that demonstrated improved performance with AMR-augmented prompting compared to context-only prompting (Llama3.1).}
\label{tab:samsum}
\begin{tabular}{|l|c|l|c|c|c|c|c|}
\hline
\textbf{Model} & \textbf{\# Examples} & \textbf{Prompt} & \textbf{Cosine similarity} & \textbf{ROUGE-1} & \textbf{ROUGE-2} & \textbf{ROUGE-L} & \textbf{BLEU} \\\hline
{\textbf{Llama3.1}} & 0 & Context-only & 66 & 29 & 8 & 21 & 2 \\
                          & 3 & Context-only & 78 & 42 & 17 & 34 & 7 \\
                          & 5 & Context-only & 78 & 43 & 17 & 34 & 7 \\
                          & LoRA & Context-only & 75 & 41 & 16 & 33 & 7\\\cline{2-8}
                          & 0 & AMR-augmented & 76 & 37 & 12 & 28 & 4 \\
                          & 3 & AMR-augmented & 79 & 41 & 17 & 32 & 8 \\
                          & 5 & AMR-augmented & \textbf{79} & \textbf{43} & \textbf{18} & \textbf{34} & \textbf{9} \\
                          & LoRA & AMR-augmented & 76 & 43 & 17 & 34 & 7\\\cline{2-8}
                          & 0 & AMR-only & 60 & 28 & 5 & 19 & 1 \\
                          & 3 & AMR-only & 79 & 41 & 17 & 32 & 8 \\
                          & 5 & AMR-only & 67 & 32 & 8 & 22 & 2 \\\hline
{\textbf{Phi3}}     & 0 & Context-only & 71 & 28 & 6 & 20 & 1 \\
                          & 3 & Context-only & \textbf{72} & \textbf{31} & \textbf{8} & \textbf{23} & \textbf{2} \\\cline{2-8}
                          & 0 & AMR-augmented & 70 & 26 & 5 & 19 & 1 \\
                          & 3 & AMR-augmented & 70 & 25 & 6 & 18 & 1 \\\cline{2-8}
                          & 0 & AMR-only & 59 & 22 & 2 & 14 & 0 \\
                          & 3 & AMR-only & 55 & 18 & 2 & 12 & 0 \\\hline
{\textbf{Mistral}}  & 0 & Context-only & 76 & 36 & 12 & 27 & 5 \\
                          & 3 & Context-only & \textbf{80} & \textbf{47} & \textbf{22} & \textbf{38} & \textbf{12} \\\cline{2-8}
                          & 0 & AMR-augmented & 77 & 38 & 14 & 29 & 6 \\
                          & 3 & AMR-augmented & 77 & 45 & 20 & 36 & 11 \\\cline{2-8}
                          & 0 & AMR-only & 66 & 30 & 6 & 21 & 1 \\
                          & 3 & AMR-only & 66 & 34 & 8 & 25 & 2 \\\hline
\end{tabular}
\end{table*}

\begin{table*}
\centering
\small
\caption{SNLI entailment prediction full results. 5-shot prompting was only coducted for the best performing model (Phi3).}
\label{tab:snli}
\begin{tabular}{|l|c|l|c|c|}
\hline
\textbf{Model} & \textbf{\# Examples} & \textbf{Prompt} & \textbf{Accuracy} & \textbf{Macro-F1} \\
\hline
{\textbf{Llama3.1}} & 0 & Context-only & 47 & 8 \\
                             & 3 & Context-only & 56 & 52 \\
\cline{2-5}
                             & 0 & AMR-augmented & 42 & 15 \\
                             & 3 & AMR-augmented & \textbf{56} & \textbf{54} \\
\cline{2-5}
                             & 0 & AMR-only & 37 & 12 \\
                             & 3 & AMR-only & 44 & 39 \\
\hline
{\textbf{Phi3}}     & 0 & Context-only & 77 & 27 \\
                             & 3 & Context-only & 82 & 80 \\
                             & 5 & Context-only & \textbf{83} & \textbf{82} \\
\cline{2-5}
                             & 0 & AMR-augmented & 67 & 39 \\
                             & 3 & AMR-augmented & 79 & 79 \\
                             & 5 & AMR-augmented & 76 & 76 \\
\cline{2-5}
                             & 0 & AMR-only & 50 & 25 \\
                             & 3 & AMR-only & 52 & 52 \\
                             & 5 & AMR-only & 55 & 55 \\
\hline
{\textbf{Mistral}}  & 0 & Context-only & 48 & 50 \\
                             & 3 & Context-only & \textbf{52} & \textbf{54} \\
\cline{2-5}
                             & 0 & AMR-augmented & 49 & 50 \\
                             & 3 & AMR-augmented & 34 & 20 \\
\cline{2-5}
                             & 0 & AMR-only & 33 & 20 \\
                             & 3 & AMR-only & 33 & 12 \\
\hline
\end{tabular}
\end{table*}

\end{document}